\documentclass{article}

\usepackage[preprint, nonatbib]{neurips_2021}

\usepackage[utf8]{inputenc} 
\usepackage[T1]{fontenc} 
\usepackage{hyperref} 
\usepackage{url} 
\usepackage{booktabs} 
\usepackage{amsfonts} 
\usepackage{nicefrac} 
\usepackage{microtype} 
\usepackage{xcolor} 

\usepackage{natbib} 
\usepackage{mathtools} 
\usepackage{siunitx} 
\usepackage{booktabs} 

\usepackage{graphicx} 
\usepackage{caption} 
\usepackage{algorithm}
\usepackage{algorithmic}

\usepackage{newfile}
\usepackage{xr}
\usepackage{tcolorbox}
\usepackage{color}
\usepackage{ifthen}
\usepackage{amsfonts}
\usepackage{amsmath}
\usepackage{amssymb}
\usepackage{subcaption}
\usepackage{array}
\usepackage{ctable}
\usepackage{comment}

\newtheorem{theorem}{Theorem}
\newtheorem{ass}{Assumption}
\newtheorem{lemma}{Lemma}

\newtheorem{proposition}{Proposition}

\newcommand{\<}{\langle}
\renewcommand{\>}{\rangle}

\newcommand{\beq}{\begin{equation}}
\newcommand{\eeq}{\end{equation}}
\newcommand{\cS}{\mathcal{S}}
\newcommand{\cA}{\mathcal{A}}

\newcommand{\cL}{\mathcal{L}}
\newcommand{\cM}{\mathcal{M}}
\newcommand{\cN}{\mathcal{N}}

\newcommand{\cG}{\mathcal{G}}
\newcommand{\bE}{\mathbb{E}}
\newcommand{\bR}{\mathbb{R}}

\definecolor{darkgreen}{rgb}{0,0.5,0}
\newboolean{showcomments}
\setboolean{showcomments}{true}
\newcommand{\desmond}[1]{\ifthenelse{\boolean{showcomments}}{\textcolor{red}{(Desmond says: #1)}}{}}
\newcommand{\thien}[1]{\ifthenelse{\boolean{showcomments}}{\textcolor{red}{(Thien says: #1)}}{}}
\newcommand{\shiau}[1]{\ifthenelse{\boolean{showcomments}}{\textcolor{red}{(SH says: #1)}}{}}
\newcommand{\laura}[1]{\ifthenelse{\boolean{showcomments}}{\textcolor{red}{(Laura says: #1)}}{}}

\newcommand{\squishlist}{
 \begin{list}{$\bullet$}
 { \setlength{\itemsep}{0pt}
 \setlength{\parsep}{3pt}
 \setlength{\topsep}{3pt}
 \setlength{\partopsep}{0pt}
 \setlength{\leftmargin}{1.5em}
 \setlength{\labelwidth}{1em}
 \setlength{\labelsep}{0.5em} } }
 
\newcommand{\squishend}{
 \end{list} }

\title{Neural-Progressive Hedging: Enforcing Constraints in Reinforcement Learning with Stochastic Programming}

\author{%
Supriyo Ghosh\thanks{This work was done while the authors were with IBM Research AI, Singapore Lab.}\\
Microsoft Research, India\\
\texttt{supriyoghosh@microsoft.com} \\
\And
Laura Wynter\\
IBM Research AI, Singapore\\
\texttt{lwynter@sg.ibm.com} \\
\And
Shiau Hong Lim\\
IBM Research AI, Singapore\\
\texttt{shonglim@sg.ibm.com} \\
\And
Duc Thien Nguyen$^*$\\
Singapore Management University\\
\texttt{dtnguyen.2014@phdis.smu.edu.sg} \\
}

\begin{document}

\maketitle

\begin{abstract}
We propose a framework, called neural-progressive hedging (NP), that leverages stochastic programming during the online phase of executing a reinforcement learning (RL) policy. The goal is to ensure feasibility with respect to constraints and risk-based objectives such as conditional value-at-risk (CVaR) during the execution of the policy, using probabilistic models of the state transitions to guide policy adjustments. The framework is particularly amenable to the class of sequential resource allocation problems since feasibility with respect to typical resource constraints cannot be enforced in a scalable manner. The NP framework provides an alternative that adds modest overhead during the online phase.
Experimental results demonstrate the efficacy of the NP framework on two continuous real-world tasks: (i) the portfolio optimization problem with liquidity constraints for financial planning, characterized by non-stationary state distributions; and (ii) the dynamic repositioning problem in bike sharing systems, that embodies the class of supply-demand matching problems. We show that the NP framework produces policies that are better than deep RL and other baseline approaches, adapting to non-stationarity, whilst satisfying structural constraints and accommodating risk measures in the resulting policies. Additional benefits of the NP framework are ease of implementation and better explainability of the policies.
\end{abstract}

\section{Introduction}

Reinforcement learning (RL) experienced a surge in popularity when deep models demonstrated superior performance in game playing with Deep Q-learning Networks (DQN) \citep{DQN}. The role of RL was cemented when it was used to beat the reigning Go world champion \citep{alphago}. Improvements to deep RL algorithms have abounded, including RL for continuous state and action spaces, with DDPG \citep{DDPG}, TRPO \citep{TRPO}, and PPO \citep{PPO}. In spite of these advances, the dominance of RL for real-world problems has lagged. We believe that this is due to three shortcomings. 

First, RL policies cannot enforce business rules, or constraints during policy execution. Yet, often structural constraints must be respected for a policy to be implementable. Existing methods, such as constrained policy gradient \citep{CPO} or ``safe" RL methods \citep{garcia2015comprehensive} do not prevent constraint violations during policy execution. Moreover, these methods can often be difficult to train and not scalable to large problems.
Second, there is a natural trade-off between expected reward and risk. The majority of RL algorithms seek to maximize the expected return. While there have been RL algorithms that optimize for various risk measures, doing so in a scalable manner and under constraints is still challenging. Third,  the sample inefficiency of RL has posed an impediment to  solving problems where high-fidelity simulators are not available to generate sufficiently large number of sample trajectories. A hope to overcoming this  is through the judicious use of models to explore more sparingly the state and action spaces.  

We introduce a framework to address these issues for problems with continuous state and action spaces. An unconstrained RL policy is first trained offline. During the online execution phase, a stochastic program (SP) is used to re-optimize the given RL policy under constraints and risk measures over a short-term future trajectory. Once the next action is chosen, the process repeats in a rolling-horizon fashion using updated state information. We call this neural-progressive hedging (NP).
The NP method  aims to exploit the generalization ability of RL to unseen scenarios jointly with  the ability of SP to exploit models and enforce scenario-dependent constraints as well as incorporating risk measures. Since the NP framework relies on a model-based online phase, it is most useful in problem settings where closed-form models of state transitions are relatively good approximations to the true state transitions. Sequential and dynamic resource allocation problems are a key example in which the NP framework excels.

Empirically, we show that the NP method results in policies that offer substantial improvements in reward under various risk measures whilst satisfying hard constraints. Moreover, we observe that the NP policy, with its more sample-efficient initial RL policy followed by the online fine-tuning phase,  is able to outperform the fully-trained (and data-hungry) RL policy. An additional benefit of the framework is ease of implementation: it can be  implemented using  existing deep RL algorithms and off-the-shelf optimization packages. To that end, the key contributions of the paper are as follows:
\begin{enumerate}
\item We define a novel method, neural-progressive hedging, combining  stochastic programming model-based online planning  with offline, deep RL for a continuous policy that satisfies hard constraints during execution; 
\item We incorporate risk-measures such as CVaR without sacrificing  model structure or decomposition algorithm; and 
\item We demonstrate the efficacy of the NP method on the class of resource allocation models, including  two  real-world problems: (i) liquidity-constrained portfolio optimization with a CVaR objective; and (ii) dynamic repositioning in a bike sharing system, where the NP method outperforms deep RL, both constrained and unconstrained as well as other baselines.
\end{enumerate}

\section{Preliminaries}
Consider the problem of  learning a deterministic policy $\pi:  \mathcal{S}\rightarrow\mathcal{A}$ in a Markov Decision Process (MDP) given by $(\cS, \cA, p, f, \gamma, T, G)$, with continuous states $s\in\mathcal{S}$, continuous actions $x\in\mathcal{A}$, transition probability distribution $p(s_{t+1}|s_t,x_t)$, cost function $f(s_t,x_t,s_{t+1}) \in\mathbb{R}$, discount factor $\gamma\in[0,1]$, decision horizon $T$, and constraint set $G$. We allow $T=\infty$ whenever $\gamma<1$. The constraint set $G$ contains a set of $K$ additional cost functions $g_1\ldots g_K$ where $g_k(s_t,x_t,s_{t+1})\in\mathbb{R}$ and constants $\beta_1\ldots \beta_K\in\bR$.
Our constrained MDP setting follows that of \citet{Altman1999ConstrainedMD}, where we aim to solve the following
problem:
\begin{align}
  \underset{\pi}{\text{minimize  }} & \bE_\pi\left[\sum_{t=1}^T\gamma^{t-1} f(s_t,x_t,s_{t+1}) \right]
  \label{eqn_problem}\\
  \text{s.t.  }  & \bE_\pi\left[\sum_{t=1}^T\gamma^{t-1} g_k(s_t,x_t,s_{t+1}) \right] \!\! \leq \beta_k , \quad k\!=\!1\ldots K.
  \notag
\end{align}
Without loss of generality we assume a fixed initial state $s_1$.
The expectation $\bE_\pi$ is taken with respect to randomness induced by the transitions
$s_{t+1}\sim p(\cdot|s_t,x_t)$ by taking $x_t=\pi(s_t)$, for all $t$.
Problem \eqref{eqn_problem} is very challenging to solve for general MDPs with continuous states and actions.
We shall now put this constrained MDP in the context of stochastic programming (SP) from which we borrow many of the algorithmic
tools in this work.

The key assumption from SP is that all the randomness or uncertainty in the system comes from external sources.
This decoupling of randomness allows us to employ powerful optimization tools in solving the main problem.
Assume $T$ is finite and let $\xi_1\ldots \xi_T$ be random variables such that the next state $s_{t+1}$ is given
by $s_{t+1}=\tilde{p}(s_t,x_t,\xi_t)$ where $\tilde{p}$ is a deterministic function once $\xi_t$ is fixed.
We call each realization of $\xi=(\xi_1\ldots \xi_T)$ a \emph{scenario}. Given a particular scenario $\xi$
\footnote{We abuse notation slightly by using $\xi$ to refer to both the random variable and its
  particular realizations.},
one can find the best action sequence in ``hindsight'' by solving the following problem:
\begin{align}
  \underset{x=(x_1\ldots x_T)}{\text{minimize}} \quad & \tilde{f}(x,\xi) \label{eqn_problem_sub}\\
  \text{s.t.} \quad & \tilde{g}_k(x,\xi) \leq \beta_k ,\quad k=1\ldots K \notag
\end{align}
where we define
\begin{flalign}
\tilde{f}(x,\xi):=\sum_{t=1}^T\gamma^{t-1} f[s_t,x_t,\tilde{p}(s_t,x_t,\xi_t) ] \hspace{0.18in} \nonumber \\
\tilde{g}_k(x,\xi):=\sum_{t=1}^T\gamma^{t-1} g_k[s_t,x_t,\tilde{p}(s_t,x_t,\xi_t) ]. \nonumber
\end{flalign}
If, for each $\xi$, the functions $\tilde{f}$ and $\tilde{g}_k$ for all $k$ are all convex in $x$, then each scenario sub-problem can be readily solved using existing convex optimization tools. To simplify
notation, we define the constraint set $\cG(\xi):=\{x|\tilde{g}_k(x,\xi)\leq\beta_k,k=1\ldots K \}$, so
problem ~\eqref{eqn_problem_sub} can be stated simply as $\text{minimize}_{x\in\cG(\xi)}\tilde{f}(x,\xi)$.

Suppose that one starts with a finite set $\Xi$ of scenarios, with known probability distribution $q(\xi)$ where
$\sum_{\xi\in\Xi}q(\xi)=1$.
One can
solve problem~\eqref{eqn_problem_sub} for each individual $\xi\in\Xi$ to obtain a mapping $x(\cdot)$
that provides
a solution $x(\xi)=(x_1(\xi)\ldots x_T(\xi))$ for each $\xi\in\Xi$. Suppose that the
action space $\cA\subseteq\bR^n$ and $|\Xi|=N$, then $x(\cdot)\in\cA^{N\times T}\subseteq\bR^{N\times T\times n}$.
How
could we then reconcile the various $x_t(\xi)$ across all $\xi\in\Xi$, at time $t$? For the resulting
solutions to be implementable, one needs to enforce a \emph{nonanticipative} property which states that
$x_t$ must only depend on information available at time $t$. From an MDP point of view, the state $s_t$ captures
all observations available up to time $t$, represented by $\xi_1\ldots\xi_{t-1}$,
and therefore $x_t$ must only depend on these if it is to be implementable, i.e., $x_t(\xi)=x_t(\xi_1,\ldots,\xi_{t-1})$
and $x_1(\xi)$ must be the same for all $\xi$.
All solutions $x(\cdot)$ that satisfy this nonanticipative property can be expressed as:
\[
x(\xi)=(x_1,x_2(\xi_1),\ldots,x_T(\xi_1,\ldots,\xi_{T-1})),\quad\forall\xi\in\Xi. 
\]
We use $\cM$ to denote the space of all nonanticipative mappings. Define an inner product on $\cA^{N\times T}$
by $\<x(\cdot),w(\cdot)\>:=\sum_\xi q(\xi)\sum_{t=1}^T \<x_t(\xi),w_t(\xi)\>$
where $\<x_t(\xi),w_t(\xi)\>$ is the standard inner product in $\bR^n$.
Given any $\hat{x}(\cdot)\in\cA^{N\times T}$, one can find a
nonanticipative version  $x(\cdot)=P_\cM[\hat{x}(\cdot)]$ where $P_\cM$ is the orthogonal projection onto $\cM$
given by the conditional expectation
$x_t(\xi)=\bE_{\xi|\xi_1\ldots\xi_{t-1}} \hat{x}_t(\xi)$ for all $t$ and $\xi$. Note that $P_\cM$ can be computed
via simple averaging over the appropriate subsets of scenarios.

Define $\cG\subseteq\cA^{N\times T}$ such that $x(\cdot)\in\cG$ iff $x(\xi)\in\cG(\xi)$ for all $\xi$.
We then aim to solve the following global problem:
\begin{equation}\label{eqn_problem_global}
  \underset{x(\cdot)\in\cG\cap\cM}{\text{minimize}} \quad \bE_\xi\tilde{f}(x(\xi),\xi) 
\end{equation}
where $\bE_\xi\tilde{f}(x(\xi),\xi)=\sum_{\xi\in\Xi} q(\xi)\tilde{f}(x(\xi),\xi)$.
Without the constraint $x(\cdot)\in\cM$, problem~\eqref{eqn_problem_global} would in fact be separable and
could be decomposed
 into solving individual scenarios as in problem~\eqref{eqn_problem_sub}. This problem, however,
can still be solved in an iterative manner where each iteration involves solving a slightly modified version
of problem~\eqref{eqn_problem_sub} for each scenario. This ``progressive
hedging'' algorithm by~\citet{rockafellar1991scenarios}, which is an application of the proximal point
algorithm, involves keeping track of 
the solution $x^i(\cdot)$ as well as a Lagrange multiplier  $\lambda^i(\cdot)$ in each iteration $i$, until convergence. It also involves a parameter $\nu^i>0$, which may be constant for all $i$.
Each iteration involves solving the following steps:
\begin{enumerate}
\item At iteration $i$, solve the following for each scenario $\xi$:
  \begin{align}\label{eqn_step_sub}
    \hat{x}^i(\xi)\in\arg\min_{x(\xi)\in\cG(\xi)}   \tilde{f}(x(\xi),\xi)+\<\lambda^i(\xi),x(\xi)\> 
       +\frac{\nu^i}{2}\|x(\xi)-x^i(\xi)\|^2
  \end{align}
\item Compute $x^{i+1}(\cdot)=P_\cM[\hat{x}^i(\cdot)]$.
\item Update the Lagrange multiplier $\lambda^{i+1}(\cdot)=\lambda^i(\cdot)+\nu^i[\hat{x}^i(\cdot)-x^{i+1}(\cdot)]$.
\end{enumerate}
In the case where $\tilde{f}$ and $\cG$ are both convex, the algorithm is guaranteed to converge to an optimal
solution $x^*(\cdot)$ of problem~\eqref{eqn_problem_global} starting from arbitrary $x^1(\cdot)$ and $\lambda^1(\cdot)$.
Local convergence to a stationary point for nonconvex $\tilde{f}$ was shown by  \citet{ Rockafellar2018ProgressiveDO}. 

The SP framework can be  adapted to  measures of risk.
Consider  CVaR,  the conditional value-at-risk, a popular  measure for finding
risk-averse solutions. The CVaR of a random variable $Z$ at
level $\alpha\in[0,1)$ can be written as:
\begin{align}
\mathrm{CVaR}_\alpha(Z):=\min_{y\in\bR}\left\{y+\frac{1}{1-\alpha}\bE_Z\left[\max\{0,Z-y\} \right]  \right\}. \nonumber
\end{align}
CVaR at $\alpha=0$ gives the expectation.
We  solve the CVaR version of problem~\eqref{eqn_problem_global}, replacing the expectation $\bE_\xi$ with
$\mathrm{CVaR}_\alpha$, by following a modified
progressive hedging algorithm \citep{rockCVAR}
with an introduction of an additional variable $y^i(\xi)\in\bR$ and the corresponding
dual $u^i(\xi)\in\bR$ for each $\xi$. Instead of equation~\eqref{eqn_step_sub}, we  solve equation ~\eqref{eqn_step_sub_cvar} in step 1 with corresponding changes in steps 2 and 3.
\begin{align}
  (\hat{y}^i(\xi),\hat{x}^i(\xi))\in & \arg\min_{y(\xi)\in\bR,x(\xi)\in\cG(\xi)} \Big\{
  y(\xi)+ \frac{1}{1-\alpha} \cdot 
  \max\{0,\tilde{f}(x(\xi),\xi)-y(\xi)\}+  \nonumber \\
 &  \frac{\nu^i}{2}|y(\xi)-y^i(\xi)|^2 + u^i(\xi)y(\xi) +\<\lambda^i(\xi),x(\xi)\>
       +\frac{\nu^i}{2}\|x(\xi)-x^i(\xi)\|^2  \Big\} \label{eqn_step_sub_cvar}
\end{align}

\section{Neural-Progressive Hedging}

We introduce  Neural-Progressive Hedging (NP) method combining the generalization capability of offline RL with the ability of SP through an online  phase to exploit models while enforcing scenario-dependent constraints and risk measures. The key steps of the NP method are shown compactly in Algorithm~\ref{neuralprox}.

The NP method works as follows:  an unconstrained RL policy $\pi_{\theta}$, parameterized by $\theta$, is obtained by solving~\eqref{eqn_problem}, or its risk-aware counterpart, without  constraints. 
In each time-step $\tau$, the NP method observes  current state $s_{(\tau)}$ and queries  RL policy $\pi_{\theta}$ to get  initial action $x^{\pi}(\cdot)$. The new NP policy is  guided by the initial RL policy via a convex combination parameter $\kappa$ so that, at convergence, the executed actions satisfy   constraints of $\cG$ and  the risk measures. Inner iterations  are denoted by $i=1, \ldots$; at each iteration $i$,  the SP sub-problems are solved for each scenario $\xi\in\Xi$ with updated Lagrangian multipliers $\lambda^i, u^i$ to obtain the dual solution $\hat{x}^i(\xi)$ and $\hat{y}^i(\xi)$. Then, we project $\hat{x}^i(\xi)$ onto a feasible space $P_{\cM}[\hat{x}^i(\cdot)]$, that satisfies the \emph{nonanticipative} property, by averaging over all the scenarios. The primal solution $x^{i+1}(\cdot)$ is obtained as a convex combination with the initial RL policy $x^{\pi}(\cdot)$ then projected with $P_{\cM}[\hat{x}^i(\cdot)]$. We then update  multipliers $\lambda^i, u^i$,  and parameters $\kappa^i$, and   $\nu^i$. This iterative process continues until the difference between  primal and dual solutions, $\delta^i$, is below a pre-defined threshold  $\epsilon$.

\begin{algorithm}[tb]
 \caption{Neural-Progressive   Hedging Algorithm }
 \label{neuralprox}
\begin{algorithmic}
	\STATE \textbf{Initialization:}  Obtain RL policy $\pi_\theta$. Define inner convergence criterion $\epsilon$, convex combination parameters $\kappa^i$ and  penalty parameters $\nu^i >0$  for $i>0$. 
	\FOR{$\tau = 1,2,\ldots$,  }
		\STATE   Observe state $s_{(\tau)}$. Sample scenario set $\Xi$, and query $\pi_\theta$ to obtain $x^\pi(\cdot)$. Set $x^1(\cdot)=x^\pi(\cdot)$. Set $\lambda^1(\cdot)=0$, $u^1(\cdot)=0$ and $i=1$.  
		\WHILE{ convergence criterion $\delta^i > \epsilon$ }
			\STATE  \textbf{1. }Solve,  for each $\xi\in\Xi$, \eqref{eqn_step_sub_cvar} (or~\eqref{eqn_step_sub} for the risk-neutral case) to obtain $\hat{x}^i(\xi)$ and $\hat{y}^i(\xi)$.
			\STATE \textbf{2. }Set $x^{i+1}(\cdot) = \kappa^i  x^\pi(\cdot) + (1-\kappa^i)  P_{\cM}[\hat{x}^i(\cdot)]$. Set   ${y}^{i+1}(\cdot)  = \bE_\xi[\hat{y}^i(\cdot)].$
			\STATE \textbf{3. }Update  multipliers:  $\lambda^{i+1}(\cdot) = \lambda^i(\cdot) + \nu^i (\hat{x}^i(\cdot)-x^{i+1}(\cdot)) $ and $u^{i+1}(\cdot) = u^i(\cdot) + \nu^i (\hat{y}^i(\cdot)-y^{i+1}(\cdot)).  $ 
			\STATE \textbf{4. }Update $\kappa^i$,  $\nu^i$.
			\STATE \textbf{5. }Convergence test:  $\delta^{i+1} := \| \hat{x}^i(\cdot)  - x^{i}(\cdot) \| + \| \hat{y}^i(\cdot)  - y^{i}(\cdot) \| $
			\STATE \textbf{6. }Set $i\leftarrow i+1$, continue.
		\ENDWHILE
		\STATE From converged solution $x^*(\cdot)$, obtain and execute $x^*_1$.
	\ENDFOR
\end{algorithmic}
\end{algorithm}

\paragraph{Resource Allocation Problems:}
The NP approach is particularly
effective for the class of resource allocation problems. In such applications, the main source of
uncertainty is  external -- consider stock price changes
or customer demands -- and to a large extent such random variables are unaffected by the
actions of the policy. A scenario generator  can hence be readily trained using  historical data.
The  set of scenarios, $\Xi$,  is obtained by sampling from such a scenario generator.
Given a scenario, $\xi$,
this policy can then be queried at any state $s_t$ to obtain the corresponding action $x_t$. Given a finite scenario
set $\Xi$, we can  obtain from $\pi_\theta$ its solution $x^\pi(\cdot)\in\cM$.

\subsection{Theoretical Analysis}
The parameter $\kappa^i$  blends the offline RL policy
with the solution from SP (Step 2 in Algorithm~\ref{neuralprox}). The assumption below covers the settings of warm start, where $\kappa^1=1$ and $\hat{\imath}=2$, and imitation learning, where $\kappa^i$ is a decreasing sequence such as $(1+i)^{-2}$,  where $1\leq \hat{\imath} < \infty$.
\begin{ass} [Imitation learning and warm start]
Let $\kappa^i\rightarrow 0$ as $i\rightarrow\infty$. Furthermore, there exists an $\hat{\imath}$ such that for  all $i\geq \hat{\imath}$, $\kappa^{i}=0$. 
\label{kappa}
\end{ass}

\begin{ass}[Existence and local convexity] Assume that the  solution set of  equation \eqref{eqn_step_sub_cvar} for a CVaR objective, or equation \eqref{eqn_step_sub} otherwise, is nonempty and finite, $\cG(\xi)$ is convex and compact, the gradients of $\tilde{f}$ are locally Lipschitz for each $\xi$ and that the dual penalty parameters $\nu^i$ are sufficiently large for all $i$.
\label{exist}
\end{ass}

 \begin{lemma}
 Under Assumption \ref{kappa}, the NP algorithm is equivalent to the progressive hedging algorithm over  an infinite number of iterations. \label{lemma1}
 \end{lemma}
{\bf Proof:}
     Assumption \ref{kappa} states that there exists a finite iterate $\hat{\imath}$ such that for all $i\geq\hat{\imath}$, $\kappa^i=0$. Since
      $x^{i+1}(\cdot) = \kappa^i  x^\pi(\cdot) + (1-\kappa^i)  P_{\cM}[\hat{x}^i(\cdot)]$,  for all $i'\geq\hat{\imath}$, $x^{i'}(\cdot) =  P_{\cM}[\hat{x}^{i'}(\cdot)]$, and hence the update of the primal variable  of the algorithm reduces to the progressive hedging update.
$\blacksquare$

Instances of stochastic programming typically make use of discretized support $\Xi$. We thus define the problem~\eqref{eqn_problem_global} in terms of a discrete $\Xi$ and refer to this problem for the remainder of this section.

\begin{ass}[Discrete support]
Let   $\Xi$  be a discrete support and let $1\ldots K$ index each scenario corresponding to a random variable $\xi\in\Xi$, with probability $p_k=1/K$. Then, problem (3) can  be expressed as: 
\beq
\min_{x_k\in{\cal G}_k; x_k\in {\cal M}} \frac{1}{K}\sum_{k=1\ldots K} \tilde{f}_{k}(x_k).
\label{eq:discproblem}
\eeq
\label{ass:discrete}
\end{ass}
\begin{theorem}[Convergence of Alg. \ref{neuralprox} for Convex  $\tilde{f}$ ] 
Under Assumptions~\ref{kappa}, \ref{exist} and \ref{ass:discrete}, along with the convexity of $\tilde{f}$,  the sequence of iterates  $(x^i(\cdot),y^i(\cdot), \lambda^i (\cdot), u^i (\cdot))$   generated by the NP algorithm  is such that 
\beq
\begin{aligned}
&\|x^{i+1} - x^i \|^2 + \| y^{i+1}- y^i \|^2 + (1/\nu^2) \| \lambda^{i+1} - \lambda^i \|^2 + (1/\nu^2) \|u^{i+1} - u^i\|^2  < 
\nonumber\\
&  \|x^i - x^{i-1}  \|^2 + \|y^i- y^{i-1} \|^2 +  (1/\nu^2) \| \lambda^i - \lambda^{i-1 }\|^2 + (1/\nu^2) \|u^i - u^{i-1}\|^2, \text{ and }
  \end{aligned}
 \eeq
 \beq
\begin{aligned}
& |x^{i+1} - x^*|^2 + |y^{i+1}- y^* \|^2 + (1/\nu^2) \| \lambda^{i+1} - \lambda^* \|^2 +  (1/\nu^2)  \|u^{i+1} - u^*\|^2    <  \nonumber \\
& |x^i - x^*|^2 + |y^i- y^*|^2 + (1/\nu^2) \| \lambda^i - \lambda^* \|^2 + (1/\nu^2)  \|u^i - u^*\|^2
 \end{aligned}
 \eeq
 with equality at $(x^*(\cdot), y^*)$ in the case of finite convergence, and thus converges to a  local solution  $(x^*(\cdot), y^*)$ with $(\lambda^*(\cdot), u^*(\cdot))$ as $i\rightarrow\infty$.   
 \label{conv}
\end{theorem}
{\bf Proof:}
    From Lemma 1, Algorithm 1 is equivalent to the Progressive Hedging Algorithm of \citet{Rockafellar2018ProgressiveDO} when run for an infinite number of iterations. The convergence of the Progressive Hedging Algorithm to a solution $(x^*(\cdot), y^*(\cdot))$ is thus guaranteed under  Assumptions 2 and 3 along with the convexity of $\tilde{f}$. 
$\blacksquare$

\begin{theorem}[Convergence of Alg. \ref{neuralprox} for Nonconvex  $\tilde{f}$] 
Let  Assumptions~\ref{kappa}, \ref{exist} and \ref{ass:discrete}, hold and let $(x^i(\cdot),y^i(\cdot))$ be a  locally optimal solution to each subproblem \eqref{eqn_step_sub_cvar}. If  sequences $\{x^i,y^i,\lambda^i,u^i\}$ converge to point $\{x^*,y^*,\lambda^*,u^*\}$, then $(x^*(\cdot),y^*(\cdot))$ generated by the NP algorithm  is a locally optimal solution to problem \eqref{eqn_problem_global}.
\end{theorem}
{\bf Proof:}        
From Lemma 1,  Algorithm 1 is equivalent to the Progressive Hedging Algorithm of \citet{Rockafellar2018ProgressiveDO} when run for an infinite number of iterations. For nonconvex $\tilde{f}$, when the Progressive Hedging Algorithm converges to a point, under Assumptions 2 and 3, it was shown in \citet{rockafellar1991scenarios} that the point is   a stationary point of the problem (3).
$\blacksquare$

The NP algorithm uses a decomposition of the measurability constraints on the scenario tree from the scenario-specific constraints, and then proceeds to solve the SP by standard optimization methods. It should be noted however that the structure and theoretical properties of the NP hold equally  with  sample average approximation \citep{saa2018}.

When combining the unconstrained policy $ x^\pi(\cdot) $ with the constrained solution $ P_{\cM}(\hat{x}^i(\cdot)) $, we also show how the quality of $ x^{i+ 1} $ evolves as a function of  $ x^\pi(\cdot) $ and $ P_{\cM}(\hat{x}^i(\cdot)) $.
\begin{proposition}
	Let $ \tilde{f}$ be Lipschitz  $\forall \xi$, i.e., $ \|\tilde{f}(x(\xi),\xi) - \tilde{f}(x'(\xi),\xi) \|\le L\|x(\xi) - x'(\xi)\| $. We have the following bound as a function of $\kappa^i$ and  Lipschitz constant $L$:
	\beq
	\begin{aligned}
	 \bE [ \tilde{f}(x^{i + 1} (\cdot), \cdot) ] \leq \bE[ \tilde{f}(x^\pi (\cdot), \cdot) ] + L(1-\kappa^i) \cdot  \|  P_{\cM}(\hat{x}^i(\cdot)) - x^\pi(\cdot)\|.\label{eq:SP_bound}
	\end{aligned} 
	\eeq
\end{proposition}
{\bf Proof:}  
	For each scenario $ \xi $, we have 
	\begin{flalign}
		\tilde{f}(x^{i + 1} (\xi), \xi) - \tilde{f}(x^\pi (\xi), \xi) & \le  L\|x^{i + 1} (\xi) - x^\pi (\xi)\| 
		 \le  L \| \kappa^i    x^\pi(\xi) +  \nonumber \\ 
		& (1-\kappa^i) \cdot P_{\cM}(\hat{x}^i(\xi)) -  x^\pi(\xi)\| \le L(1-\kappa^i) \|  P_{\cM}(\hat{x}^i(\xi)) - x^\pi(\xi)\| \blacksquare \nonumber
	\end{flalign}

Naturally, we expect an unconstrained RL solution to achieve a higher objective value, but the executed solution may include constraint violations and excessive risk. The parameter $\kappa$, thus controls the trade-off between a higher objective value and constraint  satisfaction and risk aversion.

\section{Experimental Results}
To evaluate the performance of the proposed neural-progressive hedging (NP) approach, we conduct experiments on two real-world domains where risk measures and constraints are an integral part of implementable policies:
(i) \emph{Liquidity management through portfolio optimization} which seeks to optimally reinvest earnings based on the CVaR whilst maintaining sufficient liquidity; and 
(ii) \emph{Online repositioning} which seeks to dynamically match supply-demand   when resources (here, represented by  bikes in a bike-sharing system) must be continuously rebalanced to meet  changes in demand  whilst respecting the station capacity constraints.

We compare  performance of  NP method with Constrained Policy Optimization (CPO) \citep{CPO}, and   Lagrangian-relaxed Proximal Policy Optimization (PPO-L) \citep{PPOL}. 
 DDPG \citep{DDPG} is used to solve the unconstrained RL problems. Note that when $\kappa=1$, the NP approach returns the DDPG solution. Similarly, when $\kappa=0$, the NP method returns the results of a  pure stochastic program (SP), computed using progressive hedging method \citep{Rockafellar2018ProgressiveDO}.
 
\paragraph{Experiment settings:}
We perform all the experiments on Ubuntu 18.04 virtual machines with 32-core CPU, 64 GB of RAM, and a single Nvidia Tesla P100 GPU. The distributed Ray framework and RLlib \citep{liang2017ray} were used for the DDPG method. The pure SP and NP methods with linear and non-linear objective function are solved using IBM ILOG CPLEX 12.9 and IPOPT \citep{wachter2006implementation}, respectively. The CPO and PPO-L methods are solved using OpenAI safe RL implementation \citep{PPOL}.

The unconstrained RL policy used as an expert is computed at each time step $t$ using  the DDPG algorithm \citep{DDPG}. We use a recurrent neural network (RNN) architecture for training the DDPG method with 1 hidden layer consisting of 25 hidden predictor nodes and a tanh nonlinear activation function. In addition, a long short-term memory (LSTM)  model is used to represent the RNN architecture with LSTM cell size 256 and maximum sequence length of 20.
Parameter values are as follows: the discounting factor $\gamma = 0.99$, minibatch size $b = 50$ and learning rate $lr=3e^{-5}$.  
For both constrained RL methods (i.e., CPO and PPO-L), we use a neural network with 2 hidden layers, each consisting of 256 hidden nodes with \emph{tanh} nonlinear activation function\footnote{The source codes for the CPO and PPO-L can be found at: https://github.com/openai/safety-starter-agents.}.

A discretized scenario tree is used in each decision epoch to solve the NP method for the experiments. For the financial planning example, in each decision period $t$, we generate a two layer scenario tree where the first layer consists of a root node and  the second layer  includes 1000 nodes, giving rise to 1000 scenarios. The interest rates for each of the scenarios are sampled from a multi-dimensional log normal distribution whose mean and covariance matrix are estimated from the training data set of price movements in the S\&P500.
For the  liquidity constraints, we sample 10 liquidity demand processes from a Gaussian distribution with $\mu=0.025$ and $\sigma=0.01$, giving rise to 10,000 scenarios in the second layer of the scenario tree. 
For the bike sharing problem, due to its complex non-linear objective function, we generate a two-layer tree with 200 leaf nodes, giving rise to 200 scenarios. The demand values at stations for each of the scenarios are sampled from a multi-variate normal distribution whose mean and covariance matrix are learnt from 60 days of training demand data \citep{ghosh2019improving}. Additional implementation details regarding discretization of the pure SP and NP scenario tree are provided in Appendix \ref{implement}.

\subsection{Liquidity-constrained portfolio optimization}
The liquidity management problem seeks to optimally reinvest earnings in a portfolio based on the CVaR whilst maintaining sufficient liquidity. Too much liquidity means loss of potential returns and too little incurs  borrowing costs. Model-based forecasts of the price movements and liquidity process are generally available in practice. The overall problem thus involves computing allocations across a universe of financial instruments, given observed rewards, prices, and model-based forecasts of the price  and liquidity processes. 
We have one risk-free liquid instrument. In each time step a constraint requires that the  amount in the liquid account to satisfy forecasted demand. We consider four   portfolios, each  with nine stocks and one risk-free instrument. The state at time $t$ includes the current  allocation, observed price changes and liquidity demands up to time $t$. The action is a vector, $x_t = (x_{t,1},...,x_{t,J}),$  of  allocations across $J$ instruments at  time $t$, where $j=1$  is the liquid asset. Let $\xi_{t,1}$ be the cumulative liquidity requirement and $W_t$  the  wealth at the beginning of time $t$. The constraint set is  $\cG(\xi):=\{x |  W_t \cdot x_{t,1} \geq \xi_{t,1},  \sum_j x_{t,j} = 1, t=1\ldots T \}$. The liquidity requirement $\ell^t(\xi)$ for time $t$ is sampled from a Gaussian   $\ell\sim \cN(\mu, \sigma); \mu_{\ell}=0.025, \sigma_{\ell}=0.01$ and   accumulates over  time, i.e., $\xi_{t,1} = L^{t-1}+\ell^t $, where $L^{t-1}$ denotes the accumulated realized liquidity requirement.

We use 11 years of S\&P500 daily  data from  2009--2019. The data from 2009--2016 is used for training the unconstrained RL policy $\pi_{\theta}$ and price movement model. For  the SP and NP, in each time step, we sample 1000 scenarios from a multi-variate log-normal distribution whose parameters are learnt from the training data. Hyperparameter tuning of $\pi_{\theta}$ is done using data of 2017--2018. Tests are done on two consecutive 30 working day periods in 2019 (Jan 1-Feb 11, and Feb 12-Mar 25). It should be noted that the experiments for these two testing datasets are done independently, where we assume that the initial investment starts with 1 unit of liquid asset at the first day.

\begin{figure*}[!htb]
	\centering
	\begin{subfigure}{0.3\textwidth}
		\includegraphics[width=\textwidth]{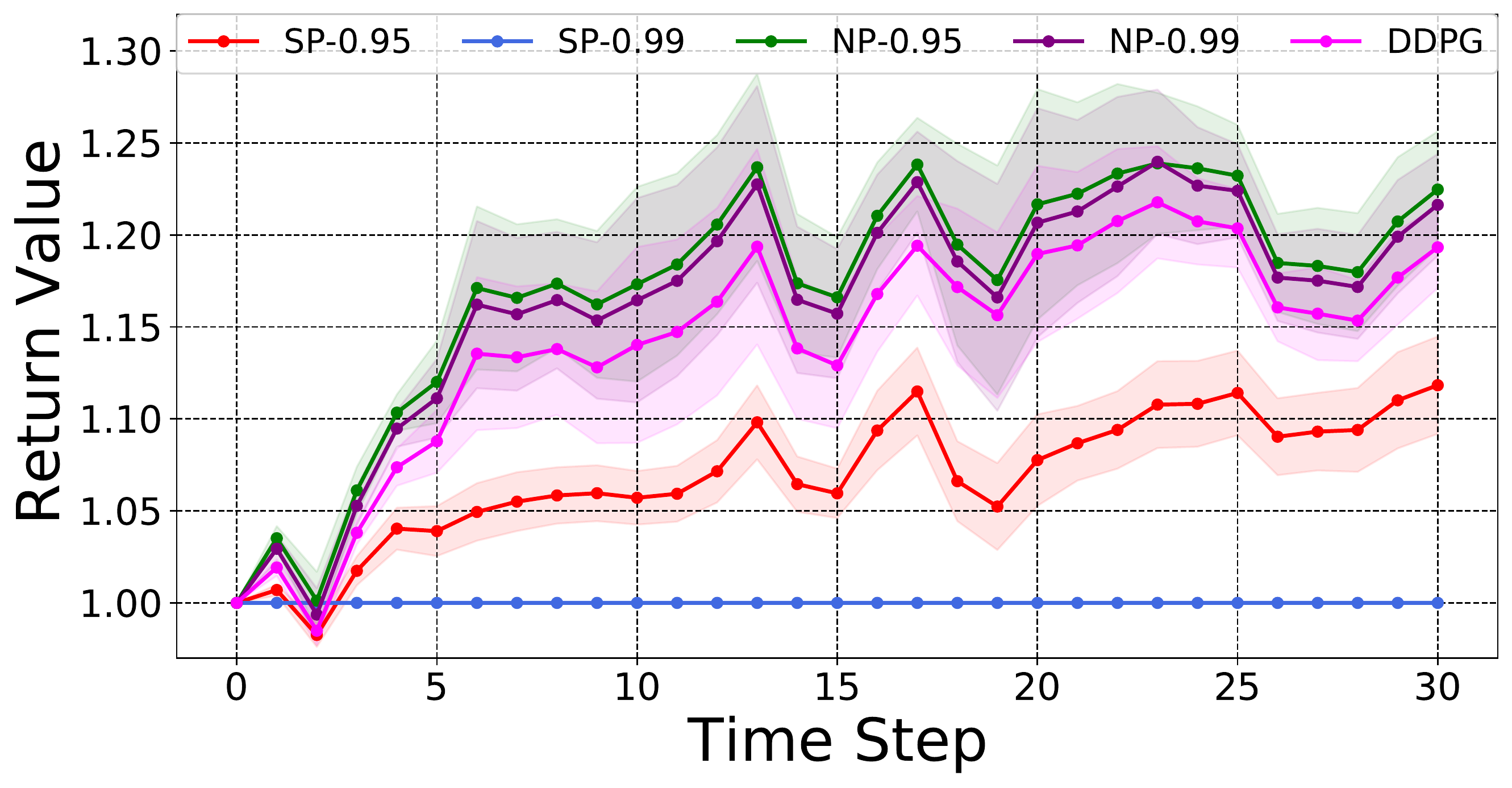} \caption{}
	\end{subfigure}  \hskip 0.2cm
	\begin{subfigure}{0.3\textwidth}
		\includegraphics[width=\textwidth]{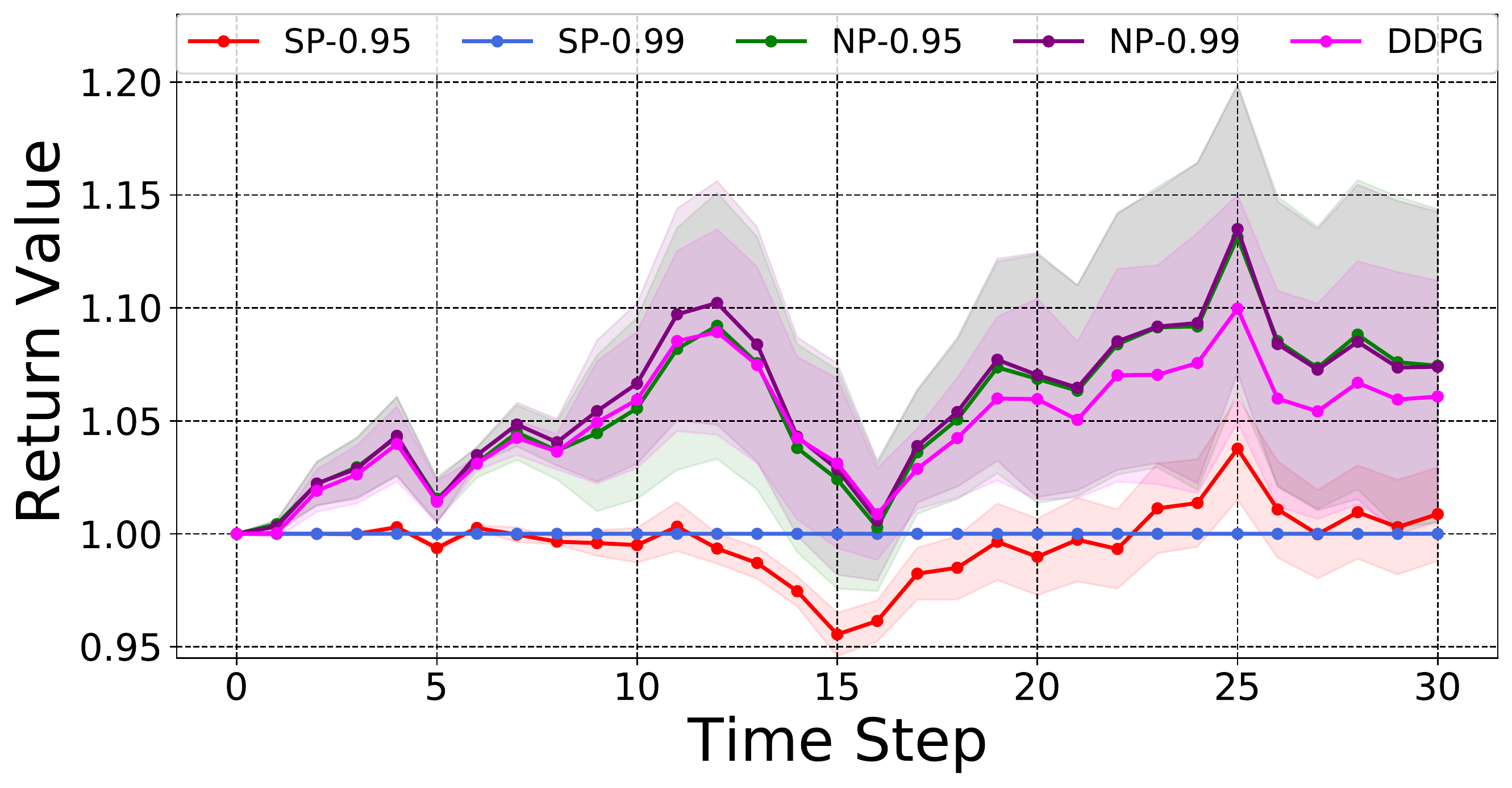} \caption{}
	\end{subfigure} \hskip 0.2cm
	\begin{subfigure}{0.3\textwidth}
		\includegraphics[width=\textwidth]{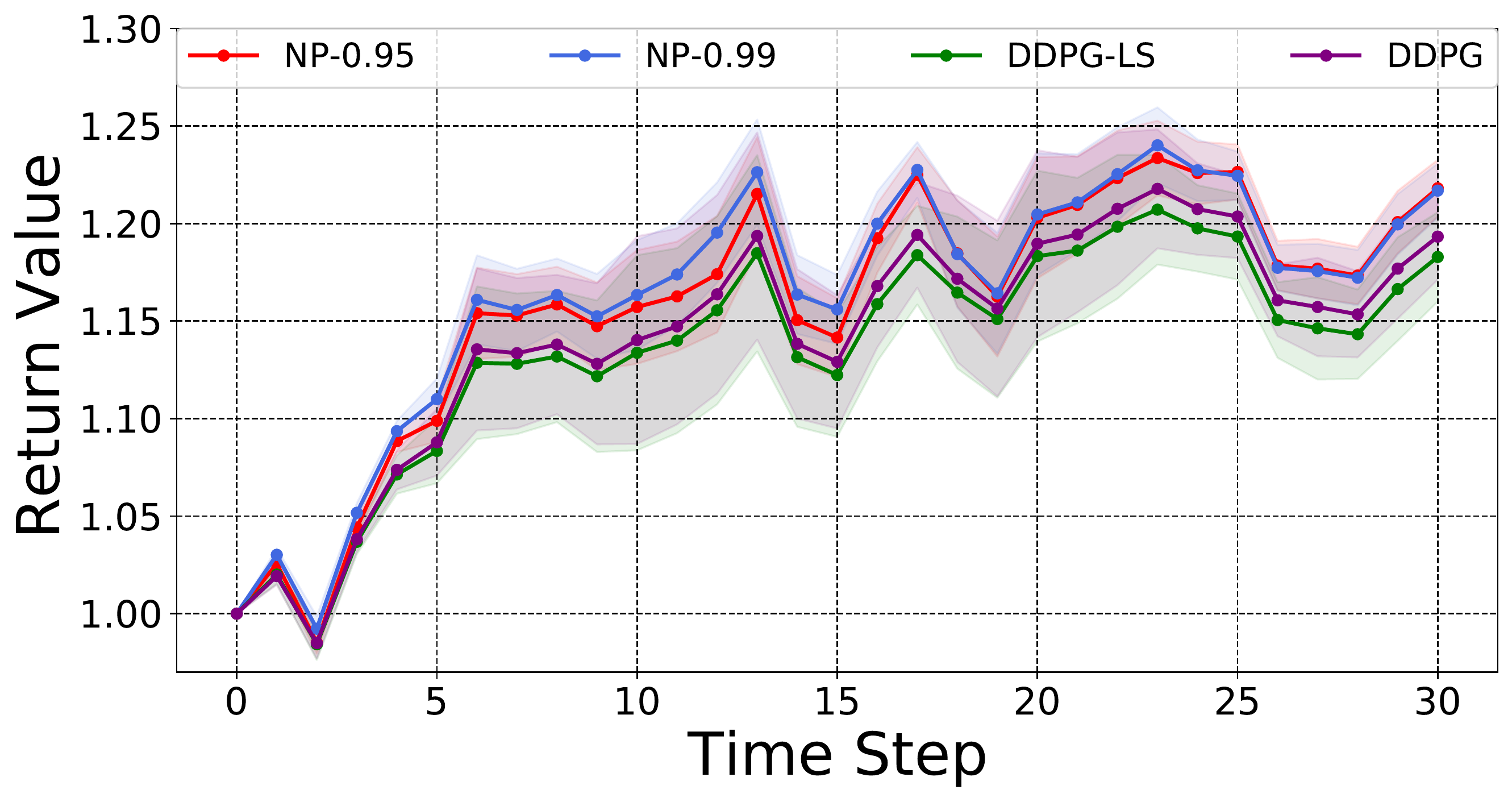} \caption{}
	\end{subfigure} 
	\caption{ (a)--(b) Returns without liquidity constraints.  NP policies  outperform   DDPG,  SP; (c) Sample efficiency of  NP.}
	\label{fig:results1}
\end{figure*}

Figure \ref{fig:results1}(a)-(b) shows the mean and standard error in returns of  NP  with CVaR  $\alpha=0.95, 0.99$, along with the unconstrained RL policy and the pure SP policy, over four asset universes.  The NP policies with CVaR  $\alpha=0.95, 0.99$ significantly outperforms the pure SP policy and improves the average return by 14\% and 18\% over the DDPG policy.
It should be noted that the variance (demonstrated by the light shaded area) arises from differences in return rates for 4 different asset universes, but our NP method always outperforms other baseline methods for individual asset universe.
Table \ref{table:metrics} provides   performance metrics  including the Sharpe ratio, volatility and maximum daily drawdown (MDD), as well as the performance of well-known baseline trading strategies. 
Specifically, we compare against four state-of-the-art online universal portfolio selection algorithms: (i) A uniform constant rebalancing portfolio (uCRP) approach \citep{cover2011universal}; (ii) Online moving average reversion (OLMAR)  \citep{li2012line} (iii) Passive-aggressive mean reversion (PAMR) \citep{PAMR} and (iv) Robust median reversion (RMR) \citep{huang2016robust}. We use a grid search to optimize the two key hyper-parameters of these universal portfolio algorithms: namely the lookback window $w$ and threshold parameter $\epsilon$\footnote{The source codes for the online portfolio selection algorithms can be found at https://github.com/Marigold/universal-portfolios.}.
Average returns and Sharpe ratios of the   NP  are higher than all the benchmark approaches. 
In Figure~\ref{fig:results1}(c), we demonstrate the sample efficiency of our expert-guided NP approach. For this experiment, we train a  DDPG policy with fewer samples (referred as ``DDPG-LS", ``LS=less samples")  obtained after 0.5 million training steps, and use it as the expert policy to guide our NP approach. Despite having less training data,  NP still provides better returns than the sample-hungry DDPG policy, which is trained to convergence at 1.5 million steps. Additional results on unconstrained portfolio optimization problem are provided in Appendix \ref{extraresults}.

\begin{table*}[!htb]
\small
\begin{center}
\begin{tabular}{>{}m{1.5cm}  >{\centering}m{1.0cm} >{\centering}m{1.2cm} >{\centering}m{1.2cm} >{\centering}m{1.0cm} >{\centering}m{1.0cm} >{\centering}m{1.2cm} >{\centering}m{1.2cm} c}
\hline
\multicolumn{1}{>{\centering}m{1.5cm}|}{} & \multicolumn{4}{>{}m{4.4cm}|}{First 30 days, annualized values} & \multicolumn{4}{>{\centering}m{4.4cm}}{Second 30 days, annualized values} \\
\hline
\multicolumn{1}{>{\centering}m{1.5cm}|} {Algorithms} &  {\footnotesize Returns  } & {\footnotesize Sharpe} & {\footnotesize Volatility} & \multicolumn{1}{>{\centering}m{1.0cm}|} {\footnotesize MDD} &  {\footnotesize Returns} & {\footnotesize Sharpe} & {\footnotesize Volatility} & {\footnotesize MDD} \\
\specialrule{.1em}{.05em}{.05em} 
{\small SP-0.0} & 11.84 & 3.58 & 27.33 & 7.63 & 2.3 & 1.16 & 17.8 & 4.8  \\
{\small SP-0.95} &  11.83 & 3.58 & 27.37 & 7.65 & 0.87 & 0.51 & 17.32 & 4.85 \\
{\small SP-0.99} &  0.0 & -0.54 & \textbf{0.0} & \textbf{ 0.0} & 0.0 & -3.78 & \textbf{0.0} & \textbf{0.0} \\
{\small \textbf {NP-0.0 }} &  \textbf{22.47} & { 4.44} & 40.29 & 10.46 & \textbf{7.44} & \textbf{2.22} & 29.1 & 7.41 \\
{\small \textbf{NP-0.95}} & \textbf{ 22.47} & {4.44} & 40.29 & 10.46 & \textbf{7.44} & \textbf{2.22} & 29.1 & 7.41 \\
{\small NP-0.99} &  21.64 & 4.29 & 40.38 & 10.44 & 7.4 & 2.09 & 30.96 & 7.99 \\
{\small DDPG} & 19.33 & 4.36 & 35.63 & 9.52 & 6.08 & 2.12 & 24.86 & 5.93\\
{\small uCRP} &  12.08 & \textbf{5.68} & 17.16 & 5.26 & 1.38 & 0.97 & 12.5 & 3.77 \\
{\small OLMAR} &  10.4 & 4.65 & 18.26 & 5.97 & -4.17 & -2.69 & 12.98 & 3.54 \\
{\small PAMR} & 6.35 & 2.45 & 22.08 & 6.03 & -8.02 & -3.39 & 20.1 & 5.19 \\
{\small RMR} &  10.68 & 4.72 & 18.45 & 5.97 & -4.56 & -2.82 & 13.58 & 3.83 \\
\specialrule{.1em}{.05em}{.05em} 
\end{tabular}
\end{center}
\caption{Performance  without liquidity constraints.  NP policies nearly always outperform all other strategies.  SP with $\alpha=0.99$ puts all funds in cash, hence MDD and volatility are  0, but returns are 0 as well.}
\label{table:metrics}
\end{table*}

\begin{figure*}[!htb]
	\centering
	\begin{subfigure}{0.3\textwidth}
		\includegraphics[width=\textwidth]{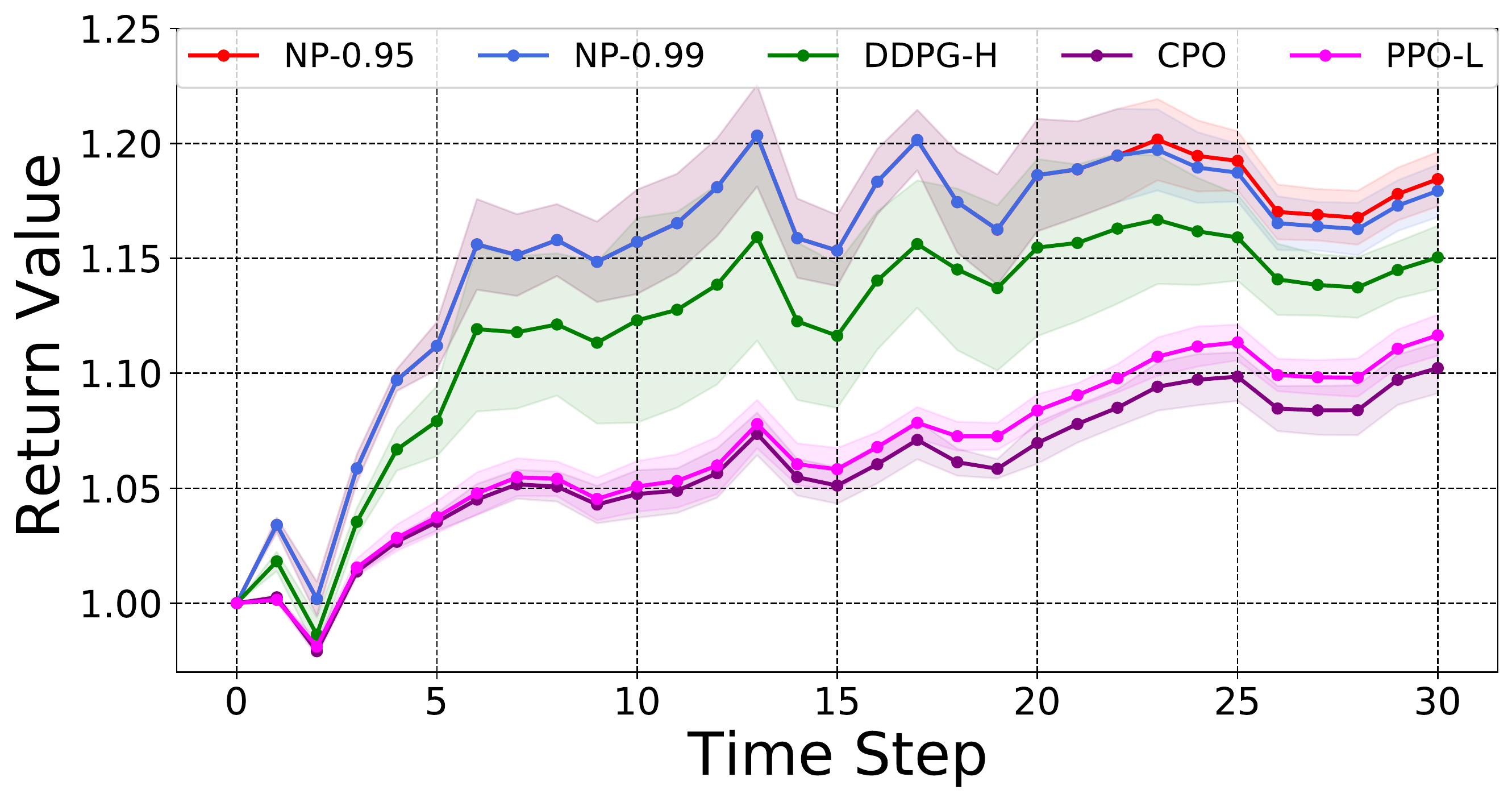} \caption{}
	\end{subfigure}  \hskip 0.2cm
	\begin{subfigure}{0.3\textwidth}
		\includegraphics[width=\textwidth]{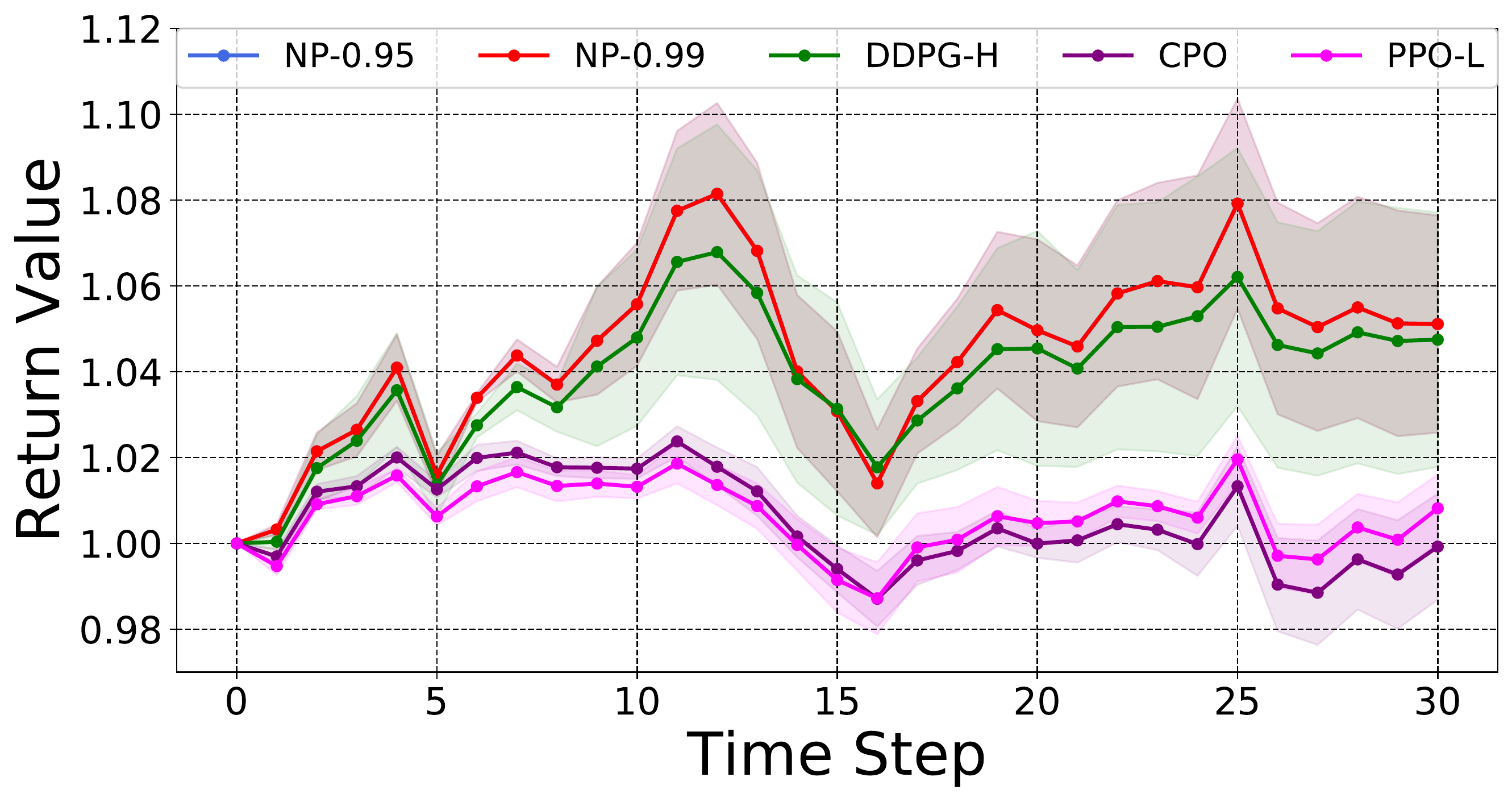} \caption{}
	\end{subfigure} \hskip 0.2cm
	\begin{subfigure}{0.3\textwidth}
		\includegraphics[width=\textwidth]{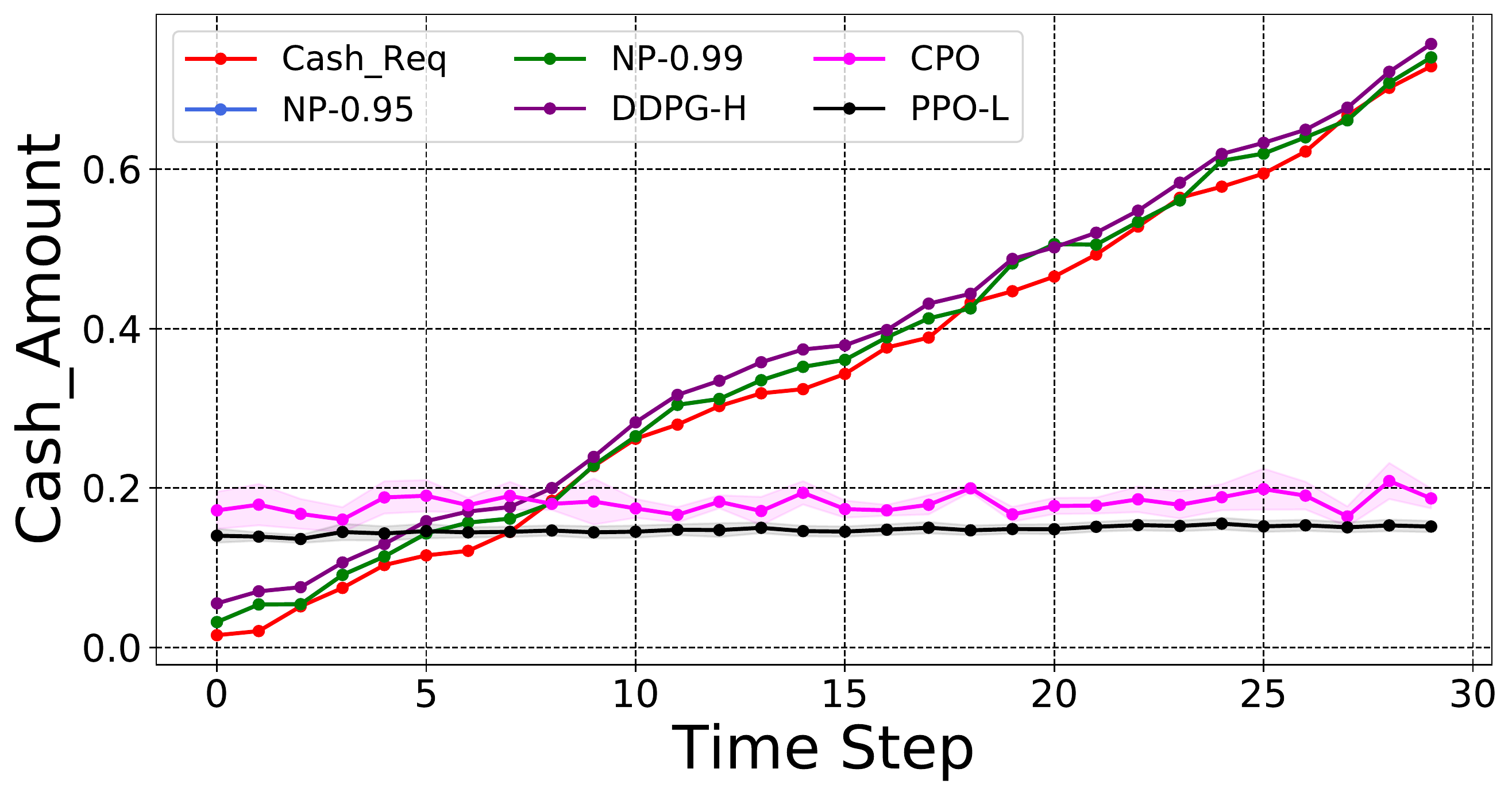} \caption{}
	\end{subfigure}  
	\caption{ (a)-(b) Returns with liquidity constraints. NP  with CVaR$_{\alpha=95}$ and CVaR$_{\alpha=99}$  are nearly identical in (b).  NP policies outperform CPO, PPO-L and DDPG-H; (c) Average liquidity in each policy, Liquidity  constraint is shown in red.}
	\label{fig:results3}
\end{figure*}
A significant benefit of the NP framework is the ability to enforce constraints, otherwise difficult to handle in an RL policy. Figure \ref{fig:results3}(a)-(b) show the mean and standard error in returns under liquidity constraints. 
We compare against a heuristic  we call DDPG-H that uses  DDPG, but  reserves $\mu_{\ell}+3\sigma_{\ell}$ of the funds  for the 0-interest cash account by re-normalizing the remaining allocations. DDPG-H thus provides  a conservative, but constraint-feasible policy, by construction.
The constrained NP policy outperforms  CPO, PPO-L, DDPG-H and even the unconstrained DDPG policy.
In Figure \ref{fig:results3}(c), we show  constraint violations; the red increasing line shows  cumulative liquidity demand in each  period. Although PPO-L, unlike CPO, was able to learn the constraints during training (see Appendix \ref{constraintViolate} for  constraint violations during training), both  CPO and PPO-L  are unable to come close to satisfying the liquidity constraints in testing. Only DDPG-H and  NP  satisfy the constraints in testing, but the  DDPG-H method over-allocates to the liquid account, thereby reducing  net returns.

\subsection{Online repositioning in bike-sharing systems}
 The bike repositioning problem is a form of online resource matching  in an uncertain environment.  Uncoordinated movements of users in  bike or electric vehicle sharing,  along with demand  uncertainty,  results in the need to often reposition  the resources \citep{ghosh2017dynamic,schuijbroek2017inventory,ghosh2016robust,ghosh2017incentivizing}. 
We use an RL-based  simulator from \citet{bhatia2019resource}  built upon the dataset of Hubway bike sharing system in Boston, consisting of 95 base stations and 760 bikes. 
The state at  time $t$ includes the current allocated bikes in each
station $j\in\{1\ldots J\}$ and $\xi_{t,j}$ is the random customer
demand at station $j$. The action is a vector, $x_t = \{x_{t,1},...,x_{t,J}\},$ that represents the percent allocations of  bikes across all  stations while respecting the constraint set $\cG(\xi):=\{x |  \check{C}_{j} \leq N \cdot x_{t,j} \leq \hat{C}_{j},  \sum_j x_{t,j} = 1, t=1\ldots T \}$, where $\check{C}_{j}$ and $\hat{C}_{j}$ denote the minimum and maximum bounds on the number of allocated bikes at station $j$, and $N$ denotes the total number of bikes present in the system. The objective function is represented by: 
\begin{align*}
\max_x \!\sum_{t} \! \sum_{j} \! & - L(x_{t,j},\xi) (1+\log (1+L(x_{t,j},\xi)) - R(x_{t,j},\xi) \sin (\pi\cdot R(x_{t,j},\xi))
\end{align*}
where $L(x_{t,j},\xi)$ represents the amount of unfulfilled demand and $R(x_{t,j},\xi)$ is the number of bikes picked up or dropped off at station $j$ at time $t$ in action $x$, which incurs a repositioning cost.
We use two months of  data to train  the models. We evaluate the learnt policies on 3 consecutive days during the morning peak period (6AM--12PM) with 12 decision epochs, each having a duration of 30 minutes.  

\begin{figure*}[!htb]
	\centering
	\begin{subfigure}{0.35\textwidth}
		\includegraphics[width=\textwidth]{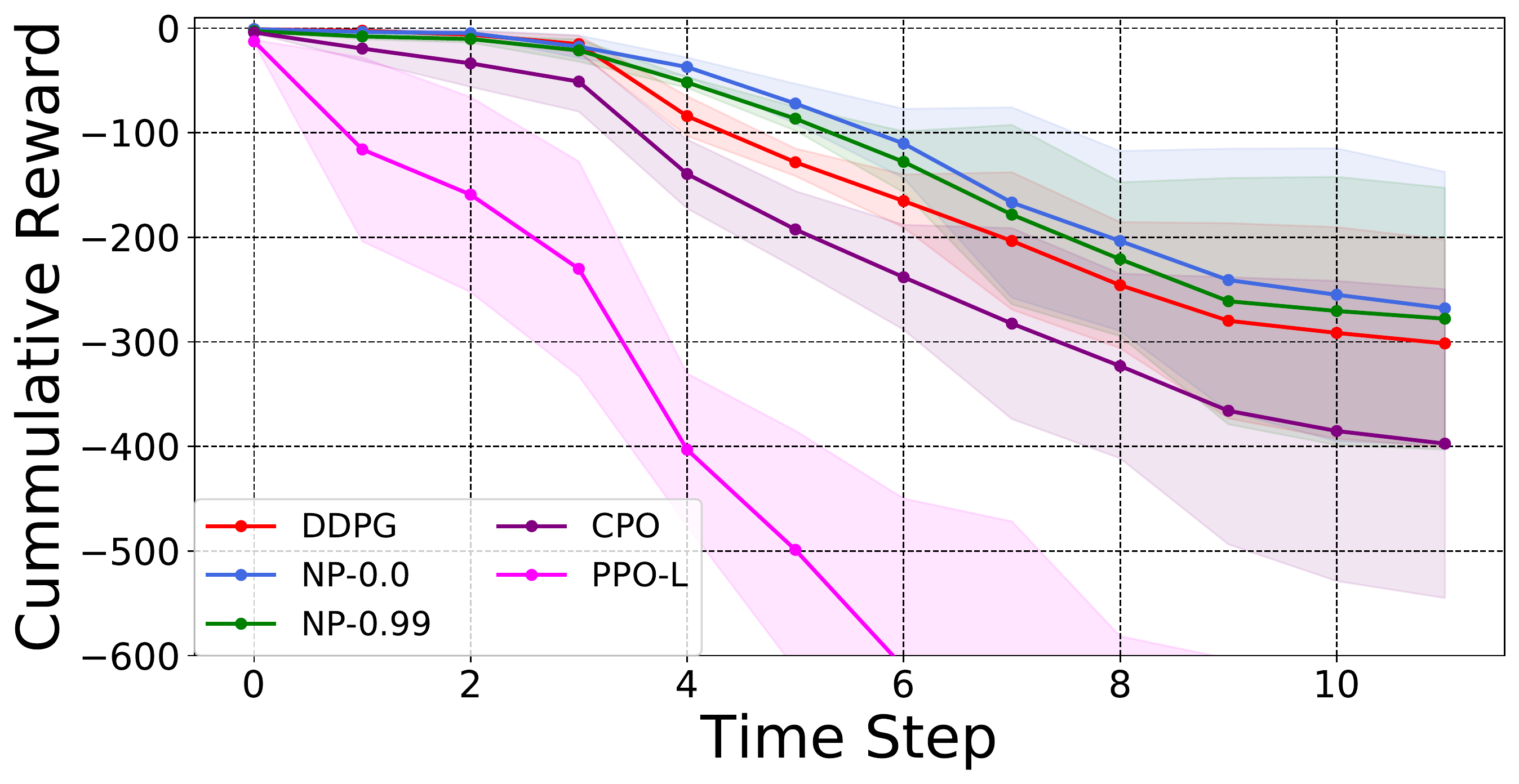} \caption{}
	\end{subfigure}  \hskip 2cm
	\begin{subfigure}{0.35\textwidth}
		\includegraphics[width=\textwidth]{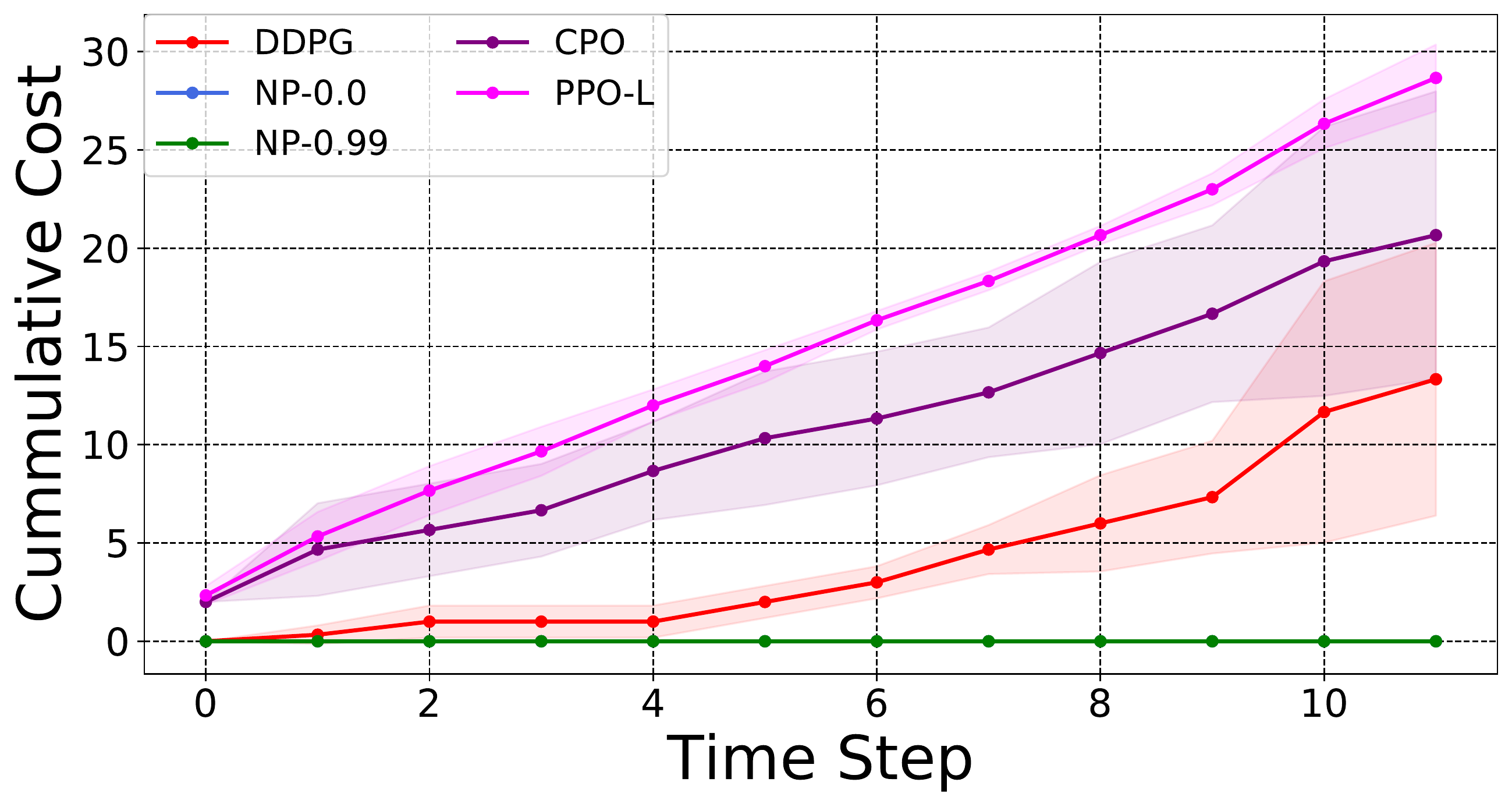} \caption{}
	\end{subfigure} \hskip .8cm
	\caption{ Performance comparison on the online bike repositioning problem for 12 time steps (6AM-12PM), averaged over 3 testing days: (a)  cumulative reward value;  (b)  cumulative constraint violation cost.}
	\label{fig:results5}
\end{figure*}

Figure \ref{fig:results5}(a) shows  mean and standard error in cumulative reward over 3 test days from NP  with CVaR of $\alpha=0.0, 0.99$,  CPO and PPO-L, and  unconstrained DDPG. After training  CPO and PPO-L methods for 1 million episodes, they fail to perform at par with  NP. NP  using expected reward (i.e., $\alpha=0.0$) and with CVAR of $\alpha=0.99$ improves  cumulative reward by 11.3\% and 7.9\% over  unconstrained DDPG. 
Figure \ref{fig:results5}(b) shows the cumulative constraint violation   where  capacity  violation  per station costs 1 unit.  Only  NP variants satisfy the constraints, while both CPO and PPO-L fail to satisfy the capacity constraints during the test period.

 \begin{figure*}[!htb]
	\centering
	\begin{subfigure}{0.35\textwidth}
		\includegraphics[width=\textwidth]{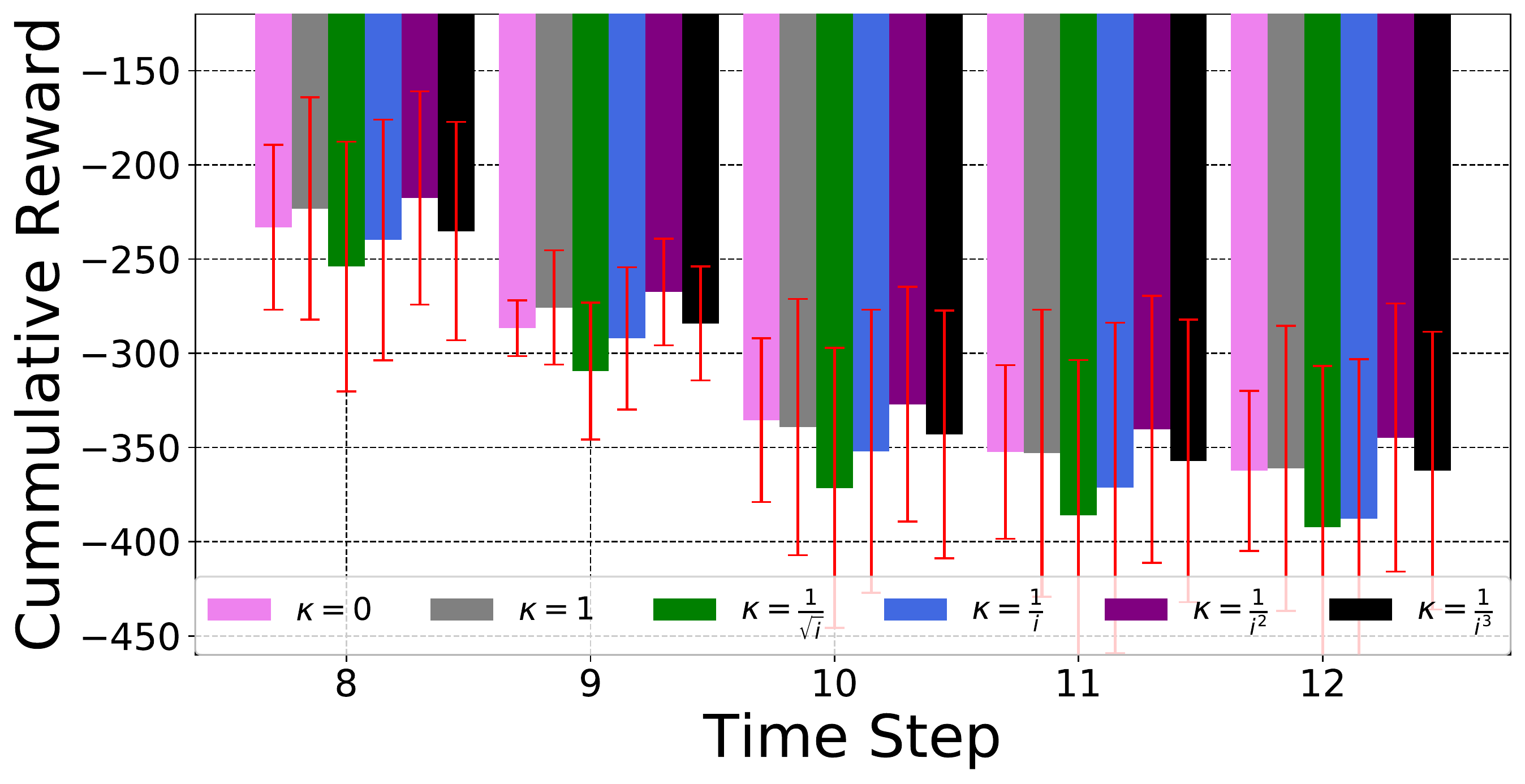} \caption{}
	\end{subfigure}  \hskip 2cm
	\begin{subfigure}{0.35\textwidth}
		\includegraphics[width=\textwidth]{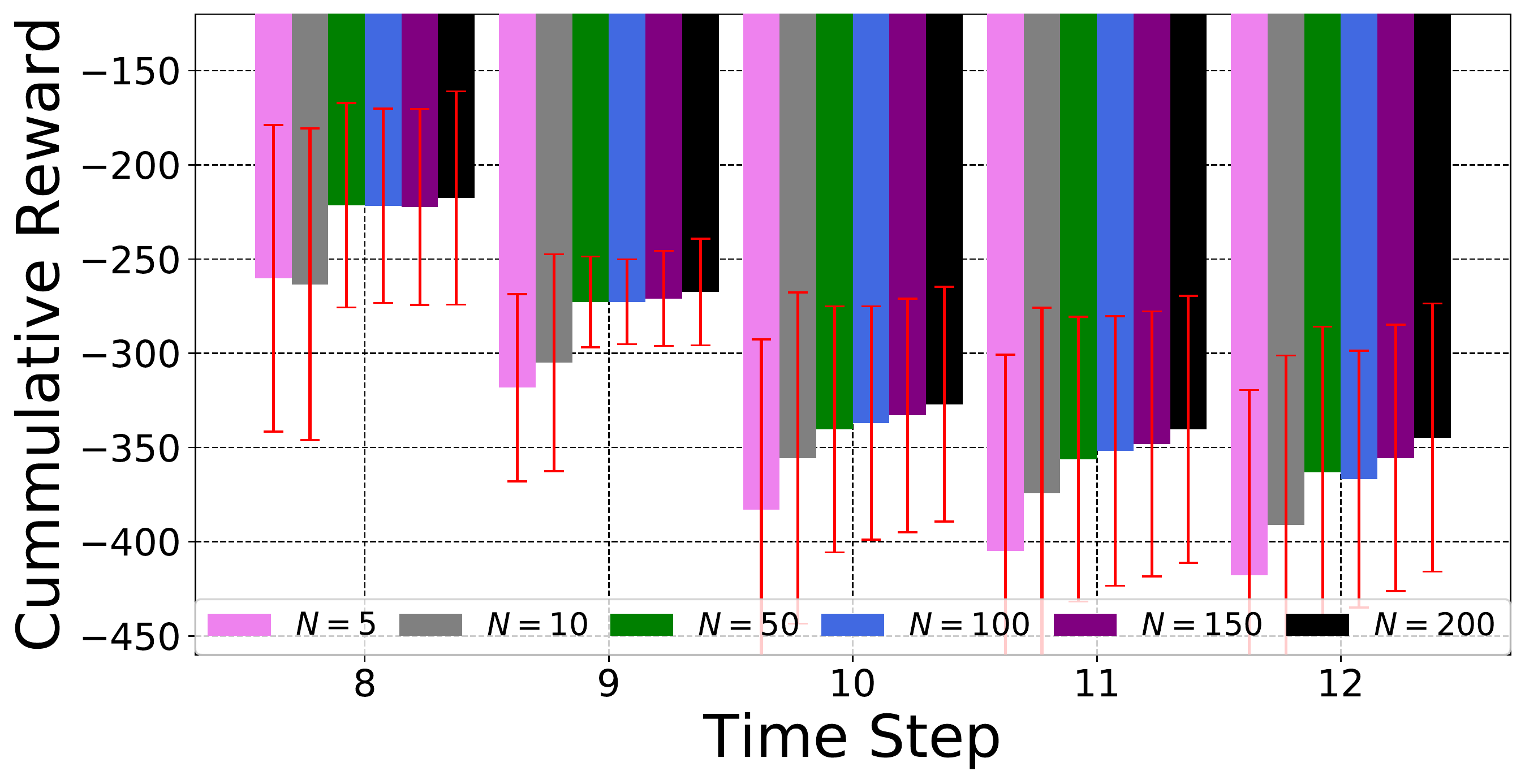} \caption{}
	\end{subfigure} \hskip 0.8cm
	\caption{ Sensitivity analysis results by varying (a)  convex combination parameter, $\kappa$; and (b)  number of scenarios.}
	\label{fig:results6}
\end{figure*}

Finally, we provide sensitivity analysis by varying (a)  convex combination parameter, $\kappa$; and (b)  number of scenarios sampled from the demand  distribution. Figure~\ref{fig:results6}(a) shows the mean and standard error in cumulative reward  for different $\kappa$  that decreases over  iterations of NP. Recall that at $\kappa=1$ and $0$,  NP  reduces  to DDPG and pure SP, respectively. The best performance is achieved with $\kappa=\frac{1}{i^2}$, which is  our default setting in the experiments. Figure~\ref{fig:results6}(b) shows the cumulative reward  of   NP  in the last five time steps, varying the number of scenarios.  NP performance improves  with the number of scenarios, but in a concave manner. We thus use 200 scenarios for both the pure SP and NP in the experiments.

\section{Related work}

The neural-progressive hedging algorithm combines offline policy search with an online model-based phase to fine-tune the policy so as to satisfy constraints and risk measures such as CVaR.
We categorize the existing relevant research into three threads: (a) Combining model-free and model-based methods for performance improvement; (b) Constrained and risk-sensitive RL methods; and (c) Improving sequential decisions through warm starting and imitation learning.

\textbf{Ensemble of model-free and model-based methods:} Model-based methods are prized for  sample efficiency, but, as noted by \citet{feinberg2018model}, high-capacity models are ``prone to over-fitting in the low-data regimes where they are most needed", implying that the combination of model-based and model-free methods will be important for good performance in complex settings. They  propose, as do \citet{nips18valueexpan}, to rollout the learned model for use in value estimation of a model-free RL, in the latter reference using an ensemble of such models to estimate variance. 
\citet{Lu2019AdaptiveOP, nips2018mpc, avivMPC2017, plato2017} combine (online) planning models with model-free RL to explore more sparingly the state and action spaces.  
\citet{ nips2018mpc} propose to differentiate through a planning model using analytical expressions for the derivatives of a  convex approximation of  a non-convex model.  
\citet{ieee2018warmstart} suggest a structure  similar to ours for controlling  dynamical systems   using models to initialize a model predictive control formulation, as a warm-start. \citet{Lu2019AdaptiveOP}  develop  Adaptive Online Planning (AOP)   with a continuous model-free RL algorithm, TF3 \citep{td3}. The goal is similar to ours -- leveraging the responsiveness of online planning with  reactive off-policy learning to make better decisions. The approach is however different from ours -- AOP uses a model-based  policy  when uncertainty is high and a reactive model-free policy when habitual behavior should suffice.

\textbf{Constrained and risk-sensitive RL methods:} \citet{garcia2015comprehensive} surveyed safe RL methods which they classify as either  optimization-based or handling safety  in the exploration process.  \citet{pham2018constrained} suggest  after each policy update to project the current iterate onto the feasible set of safety constraints; since they assume that the safety constraints may not be known in advance, they propose a method to learn the parameters of a linear polytope. \citet{yang2020} extend CPO method to solve constrained RL  by  optimizing the reward function using TRPO and then projecting the solution onto the feasible region defined by safety constraints, similar to \citet{pham2018constrained}.   \citet{chow1, chow2}  model risk-constrained MDPs with a CVaR objective or chance constraints, and solve it by relaxing the constraints and  using a policy gradient algorithm. However, similar to  CPO and  Lagrangian-relaxed  PPO, the  constraints are not enforced during execution and need not be satisfied.   Most  ``safe" RL methods use  an initial  infeasible, unconstrained policy and iteratively  render it feasible and locally optimal, e.g., \citet{berkenkamp2017safe} define an expanding ``region of attraction" to guide safe exploration to improve the policy, whilst remaining feasible.

\textbf{Imitation learning and warm start:} NP can  be viewed through the lens of  imitation learning. 
 \citet{2016continuousdeepQ}  use synthetic model-based ``imagination"  rollouts  in the early iterations of  deep RL training, which can be considered as a model-based warm-start. This is  the opposite of our approach, we propose a longer-horizon deep RL   to warm start the online stochastic program. 
Aggravate \citep{aggravate} and Aggravated \citep{aggravated}, building on the seminal DAgger \citep{dagger}, involve iteratively mixing the learning step of a policy with an expert policy, in that, at iteration $n$, $\pi^n = \beta^n \pi^* + (1-\beta^n) \hat{\pi}^n $, where  $\beta\rightarrow 0$ as $n\rightarrow \infty$.  This is similar to the update step of  NP which uses a convex combination of the expert and the learner policies, with damping. \citet{cheng2018fast} take this one step further by defining a  framework with  a mirror descent gradient update that reduces to   imitation learning-based RL, depending on the choice of the  gradient estimator; they introduce SLOLS,  where the gradient  is a convex combination of a policy gradient and an expert gradient. 
\citet{sun2018truncated} propose combining imitation learning and RL with the aim of  faster learning and improving beyond a sub-optimal expert.
 The  advantages achieved by the NP method in inverting the roles of expert and learner are the ability of the SP to enforce hard constraints and incorporate risk measures, and doing so in an explainable manner.
The NP warm start serves as an external expert to guide  the SP in the early iterations  to encourage  convergence to a better solution by reshaping the objective itself.

\section{Conclusion}
The neural-progressive hedging (NP) method  starts from an offline, unconstrained RL policy and iteratively enforces constraints and risk requirements using model-based stochastic programming. The framework is thus a type of external point method. 
We demonstrate the efficacy of NP method on two real-world applications, in finance and logistics. The NP method significantly outperforms both constrained and unconstrained RL whilst handling  both resource constraints and risk measures.
An important benefit of the framework is its ease of implementation: NP method can be implemented  using existing deep RL algorithms and commercial off-the-shelf optimization packages, and provides added transparency and explainability on the constraint satisfaction of the policy. 

\appendix

\section*{Appendix}
\section{Additional Numerical Results}
In this section, we provide additional empirical results on the convergence of NP method, the effect of warm-starting NP with an offline RL solution, and episodic constraint violation cost for CPO and PPO-L benchmarks during the training process.
\subsection{Convergence and effect of warm-starting}\label{extraresults}
Figure~{\ref{fig:results7}(a) shows the convergence of the NP method, in the presence of and after damping to zero the expert guidance. According to Theorem \ref{conv},  initial iterates may  decrease non-monotonically but for iterations $i\geq \hat{\imath}=20$  progression to an optimum is monotonic
Figure~{\ref{fig:results7}(b) compares the warm-start (called NP-WS) version with the damped-guidance, or imitation-learning-type  expert guidance (called NP). Both warm-starting and imitation learning with damping for CVaR $\alpha=0.95$ and $0.99$ perform far better than the baseline DDPG policy.

\begin{figure*}[!htb]
	\centering
	\begin{subfigure}{0.45\textwidth}
		\includegraphics[width=\textwidth]{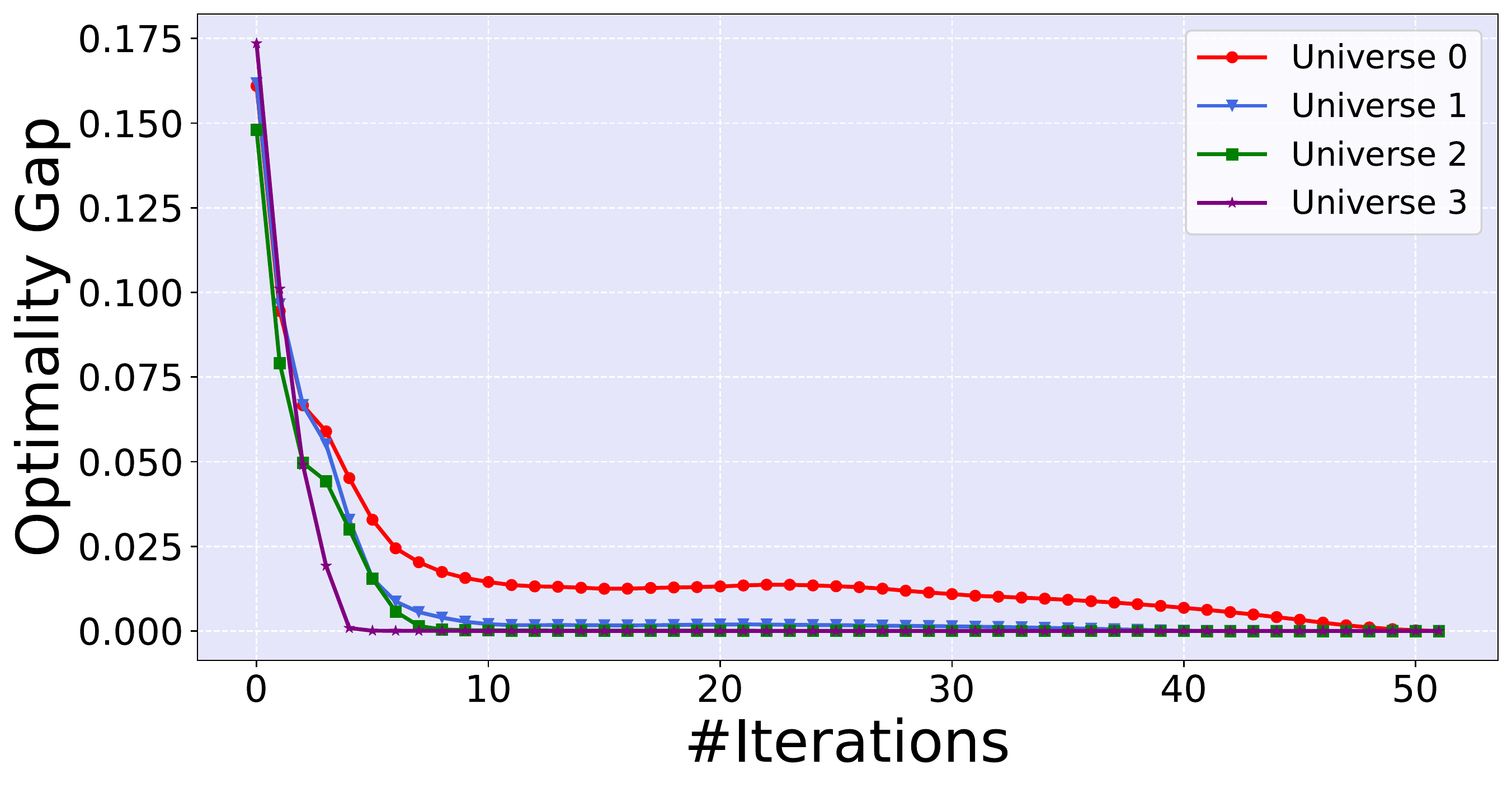} \caption{}
	\end{subfigure}  \hskip 1cm
	\begin{subfigure}{0.45\textwidth}
		\includegraphics[width=\textwidth]{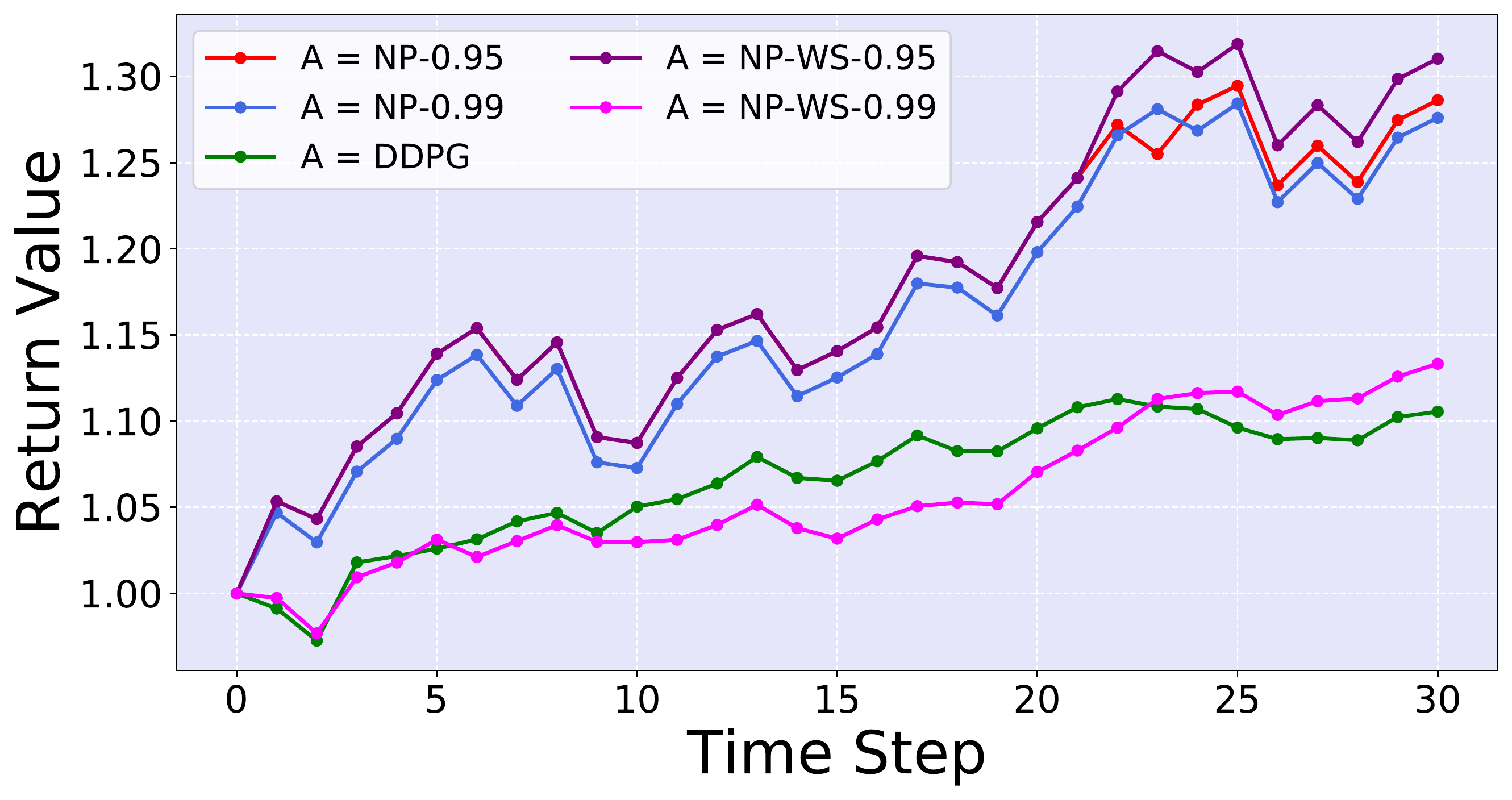} \caption{}
	\end{subfigure} 
	\caption{ (a) Convergence of the neural-progressive hedging algorithm with  CVaR$_{\alpha=95}$; (b) Performance comparison of two versions of the algorithm: warm start ($\kappa^1=1$,  $\hat{\imath}=1$) vs. imitation learning ($\kappa^i=(1+i)^{-2} $, $\hat{\imath}=20$, in this example).}
	\label{fig:results7}
\end{figure*}

\subsection{Constraint violation for benchmark constrained RL}\label{constraintViolate}
Figure~{\ref{fig:results8} illustrates the episodic constraint violation cost for two benchmark constrained RL algorithms, CPO of \citet{CPO}, and PPO-Lagrangian of \citet{PPOL}, for the financial planning domain. Each episode duration is 30 time steps and in each time step $t$, we enforce a cost of 1 if the amount available in the liquid instrument is less than the cumulative account payable up through time $t$. Observe that  CPO  fails to learn the constraints  during training. The PPO-Lagrangian method is able to bring down the episodic cost to 0 during training (the limit of the episodic cost is set to 0), but as shown in the main paper (see Figure 2(c)), the learned PPO-L policy is not able to  satisfy the constraints during execution. 

\begin{figure*}[!htb]
	\centering
	\begin{subfigure}{0.45\textwidth}
		\includegraphics[width=\textwidth]{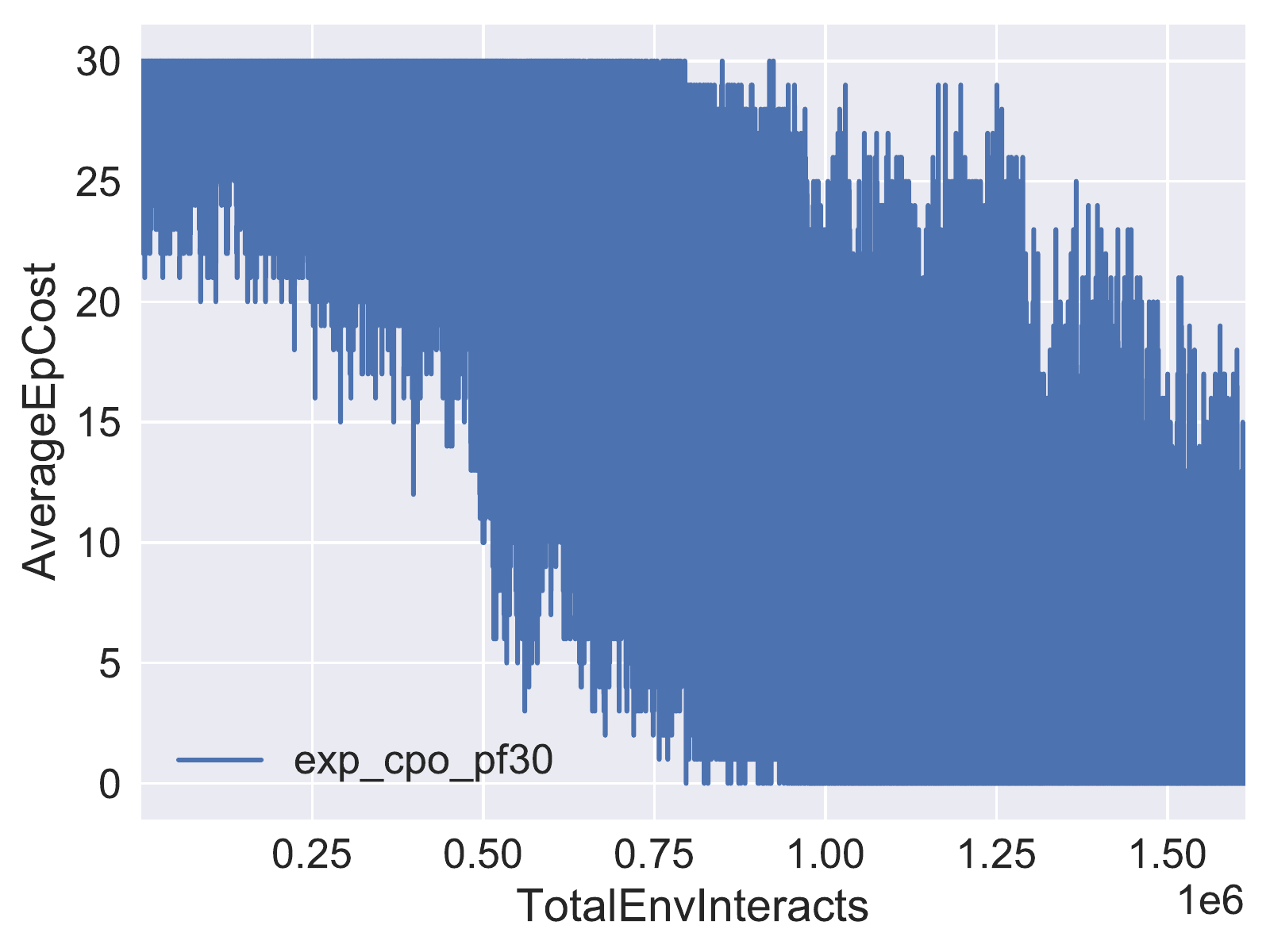} \caption{}
	\end{subfigure}  \hskip 1cm
	\begin{subfigure}{0.45\textwidth}
		\includegraphics[width=\textwidth]{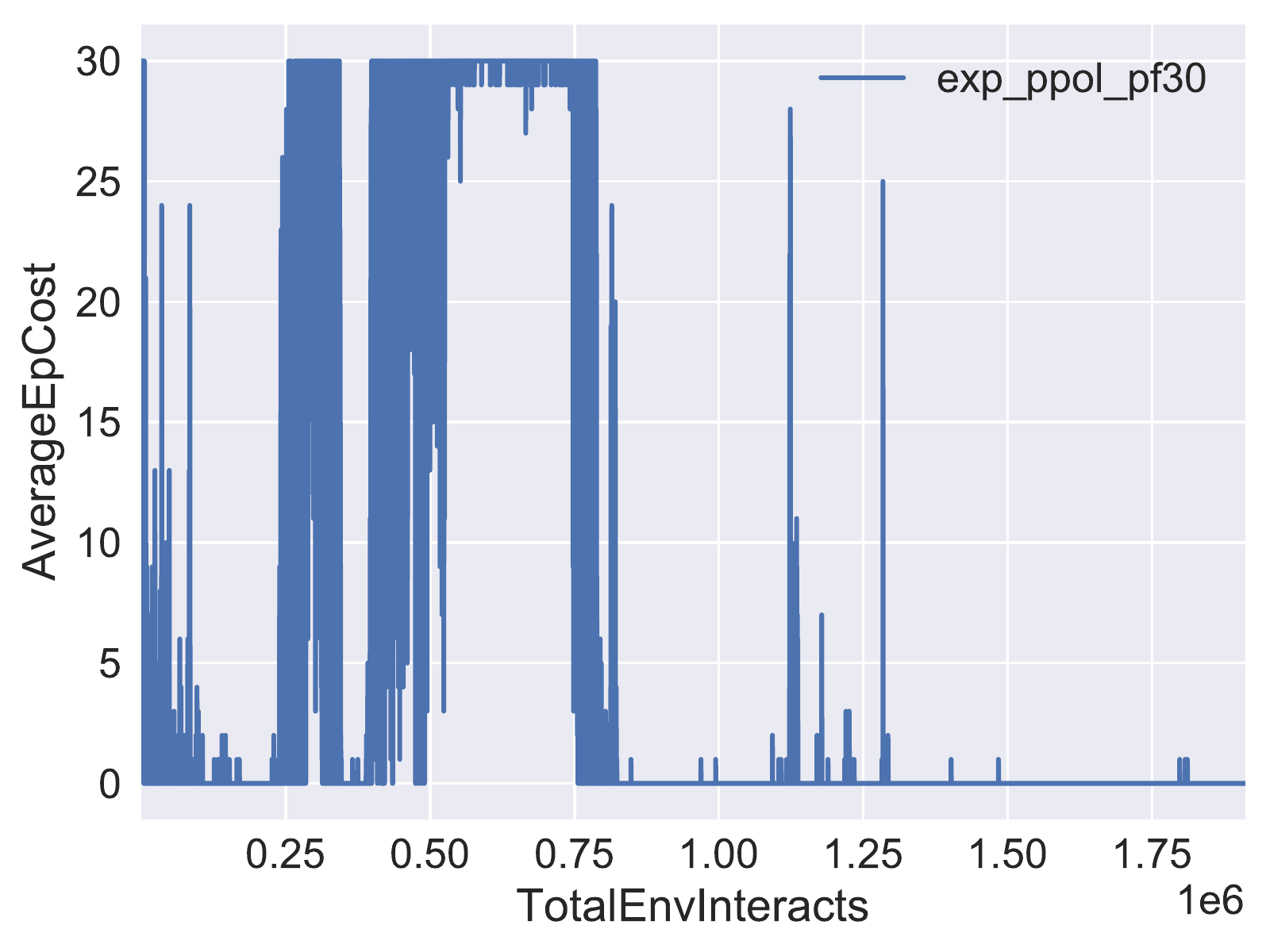} \caption{}
	\end{subfigure} 
	\caption{ Episodic (episode length = 30) constraint violation cost during training for (a) CPO; and (b) PPO-Lagrangian.}
	\label{fig:results8}
\end{figure*}

\section{Additional Implementation Details}\label{implement}

\paragraph{Discretization of the stochastic program scenario tree:}
Consider a finite scenario tree formulation of a stochastic programming problem, such that the set of nodes in the scenario tree at time stage $t$ are denoted $N_t$. 
A node denotes a point in time when a realisation of the random process becomes known and a  decision  is taken. Each node  replicates the data of the optimization problem,  conditioned on the probability of visiting that node from its parent node. A path from the root to each leaf node is referred to as a scenario; its  probability of occurrence,  $p_s$, is the product of the conditional probabilities of visiting each of the nodes on that scenario path.
The discretized model-based   stochastic program is thus:
\beq
 \max   \sum_{s=1\ldots S} F_s(x, \xi ) := \sum_{s=1\ldots  S} p_s  \sum_{t=1\ldots T} f_t(x_s(t)).
 \label{eq:SPobj2}
 \eeq 
The \textit{non-anticipativity} constraints are critical for the implementability of the policy but they couple the scenario sub-problems by requiring that the action $x_t$ at  time $t$ is the same across scenarios (i.e., sample paths) sharing the sample path up to and including time $t$. For each $\xi\in\Xi$, these coupling constraints are expressed as:
\beq
x(\xi) = ( x_1, x_2(\xi_1), x_3 (\xi_1,\xi_2), \ldots x_T (\xi_1 \ldots \xi_{T-1}).
\label{eq:meas}
\eeq 
Using the discretized formulation of \eqref{eq:SPobj2}, and
following \citet{rosa1996augmented} we can rewrite \eqref{eq:meas}  in a manner that  facilitates  relaxation of those constraints: Define the last common stage of two scenarios $s_1$ and $s_2$ as 
\beq
t^{\max}(s_1,s_2):= \max\{ \hat{t}: s_{1}(t) = s_{2}(t), t=1,\ldots \hat{t} \},
\eeq
and then re-order the scenarios $s=1\ldots S$, so that at every $s$, the scenario $s+1$ has the largest common stage with scenario $i$ for all scenarios $s' > s$, that is $t^{\max}(s,s+1):= \max\{ t^{\max}(u,v) : v>u\}$. Then, define the sibling of scenario $s$ at time stage $t$ as a permutation $\nu(s,t):= s+1$ if $t_{\max}(s,s+1)\geq t$ and $\nu(s,t):= \min\{t' : t^{\max}(s,t')\geq t \}$ otherwise. The inverse permutation shall be  denoted $\nu^{-1}(s,t).$
Note that the sibling of a scenario depends upon the time stage, and that a scenario with no shared decisions at a time stage has by definition itself as sibling. 
Using the above, \citet{rosa1996augmented}  re-define the constraints enforcing measurability in terms of the sibling function
as follows:
\beq
x_s(t) = x_{\nu(s,t)}(t) \;\; \forall (s,t), \; s \ne \nu(s,t).
\label{eq:measurability2}
\eeq
Equation \eqref{eq:measurability2} is convenient in the primal-dual formulation in terms of discrete scenarios, presented next.
We are interested in maintaining the separability of the subproblems which depend only on individual scenarios of the random variable to facilitate handling large problems via scenario-based decomposition. To do so, we relax the constraints using the following formulation
\beq
\cM := \{x:    M_1 x_1 (\xi) + \ldots + M_Sx_S  (\xi)= 0\},
\label{eq:measurability3}
\eeq
where the matrices in \eqref{eq:measurability3}  are defined so that each $M_s$ is a matrix of -1, 0 and 1 such that at the root node $x_{11}=x_{12}, x_{12}=x_{23}=\cdots x_{1,s-1}=x_{1,s}$, at the stage $t=2$, there are as many such sets of equalities as children nodes emanating from the root node, and so on up to stage $T-1$.  At stage $T$, all nodes are leaves and no such linking constraints are required. The projection of a point $x^i$ onto the subspace $\cM$, $P_{\cM}[x^i(\cdot)]$  can be computed by taking the conditional expectation of $x^i$, $E_{\xi \, | \, \xi_1, \ldots \xi_{i-1}}$.
Lagrange relaxation of the measurability constraints \eqref{eq:measurability2} gives rise to the following Lagrange function, in terms of the discrete scenarios $s=1\ldots S$:
\begin{align}
\cL(x,\lambda) = \sum_{s=1\ldots S}  p_s \sum_{t=1\ldots T} f_t(x_s(t)) + \sum_{s=1\ldots S}  \sum_{t=1\ldots T-1}  \lambda_s(t) (x_s(t) - x_{\nu(s,t)}(t)).
\label{eq:primal_dual}
\end{align}
The scenario subproblems are re-defined as a function of the inverse permutation of the sibling function:
\begin{align}
\min_{x_s \in G'_s} \cL_s(x_s,\lambda_s) = p_s \sum_{t=1\ldots T} f_t(x_s(t)) + 
\sum_{t=1\ldots T-1} ( \lambda_s(t) - \lambda_{\nu^{-1}(s,t)}(t) ) x_s(t)
\label{eq:lag1}
\end{align}
for each $s=1\ldots S$.
The dual problem is given by
\beq
\max_{\lambda} D(\lambda) := \min_{x \in  G' } \cL (x,\lambda).
\label{eq:dual1}
\eeq

It is possible to further speed up convergence of our NP algorithm in practice using the approach of \citet{zehtabian2016penalty}. This approach monitors the primal and dual gap terms in convergence criteria separately to update the penalty parameters so as to reduce the convergence gap quickly.

\bibliography{ArXiv2022NPH}
\end{document}